\documentclass[twoside]{article}

\usepackage[accepted]{aistats2024}

\usepackage{hyperref}       %
\usepackage{url}            %
\usepackage{booktabs}       %
\usepackage{hyperref}       %
\usepackage{amsfonts}       %
\usepackage{microtype}      %
\usepackage{xcolor}         %
\usepackage{amsmath}
\usepackage{graphicx}
\usepackage{caption}
\usepackage{multirow}
\usepackage[backend=biber, style=authoryear, maxcitenames=1, mincitenames=1, natbib=true]{biblatex}%
\usepackage{titlesec}
\usepackage{docmute}

\newcommand{\myparagraph}[1]{{\bf #1}\ \ }
\newcommand{\appendixref}[1]{\ref{#1}}
\newcommand{\mainornot}[2]{\ifismaindoc#1\else#2\fi}

\newcommand{\argmax}{\mathop{\rm argmax}\limits}
\newcommand{\argmin}{\mathop{\rm argmin}\limits}
\newcommand{\argtopk}{\mbox{arg top-{\it s}}}
\newcommand{\topk}{\mbox{{\it s}th-val}}
\newcommand{\topkplus}{\mbox{({\it s}+1)th-val}}
\newcommand{\mat}[1]{\boldsymbol{#1}}
\newcommand{\bmc}[1]{\boldsymbol{\mathcal{#1}}}
\newcommand{\bref}[1]{(\ref{#1})}
\newcommand{\set}[1]{\mathcal{#1}}
\newcommand{\bA}{\boldsymbol{A}}
\newcommand{\bx}{\boldsymbol{x}}
\newcommand{\by}{\boldsymbol{y}}
\newcommand{\bz}{\boldsymbol{z}}
\newcommand{\bZ}{\boldsymbol{Z}}
\newcommand{\bK}{\boldsymbol{K}}
\newcommand{\bI}{\boldsymbol{I}}
\newcommand{\bb}{\boldsymbol{b}}
\newcommand{\br}{\boldsymbol{r}}
\newcommand{\bphi}{\boldsymbol{\phi}}
\newcommand{\Del}{\mathtt{Del}}
\newcommand{\Ins}{\mathtt{Ins}}
\newcommand{\CDel}{\mathtt{CDel}}
\newcommand{\CIns}{\mathtt{CIns}}
\newcommand{\dist}{\mathtt{dist}}
\newcommand{\expand}{\mathtt{expand}}

\newif\ifismaindoc
\ismaindoctrue %

\begin{document}

\twocolumn[

\aistatstitle{Explanation-based Training with Differentiable Insertion/Deletion Metric-aware Regularizers}

\aistatsauthor{ Yuya Yoshikawa \And Tomoharu Iwata }

\aistatsaddress{ STAIR Lab, Chiba Institute of Technology \And NTT Corporation } ]

\begin{abstract}
The quality of explanations for the predictions made by complex machine learning predictors is often measured using insertion and deletion metrics, which assess the faithfulness of the explanations, i.e., how accurately the explanations reflect the predictor's behavior.
To improve the faithfulness, we propose insertion/deletion metric-aware explanation-based optimization (ID-ExpO), which optimizes differentiable predictors to improve both the insertion and deletion scores of the explanations while maintaining their predictive accuracy.
Because the original insertion and deletion metrics are non-differentiable with respect to the explanations and directly unavailable for gradient-based optimization, we extend the metrics so that they are differentiable and use them to formalize insertion and deletion metric-based regularizers.
Our experimental results on image and tabular datasets show that the deep neural network-based predictors that are fine-tuned using ID-ExpO enable popular post-hoc explainers to produce more faithful and easier-to-interpret explanations while maintaining high predictive accuracy.
The code is available at \url{https://github.com/yuyay/idexpo}.
\end{abstract}

\section{Introduction}\label{sec:intro}
Complex machine learning predictors, such as deep neural networks (DNNs), have become indispensable components of many modern AI systems because of their remarkable predictive accuracy.
In addition to having high predictive accuracy, it has been crucial in AI systems for medical diagnostics~\parencite{Holzinger2017-yt}, autonomous driving~\parencite{Omeiza2022-ud}, cybersecurity~\parencite{Capuano2022-sc}, e-commerce~\parencite{10.1561/1500000066}, etc., to explain the rationale for the predictor's behavior so that users can trust the AI system.
Such explanations also help researchers and developers to identify flaws caused by biases in training datasets~\parencite{Van_Stein2023-yl} and the errors in implementation and modeling~\parencite{lertvittayakumjorn-toni-2021-explanation}.

To understand the behavior of a predictor, it is crucial to know what features are essential to the individual predictions produced by the predictor and to what extent they are essential.
To obtain the explanations, researchers and practitioners rely on post-hoc explainers or use inherently interpretable predictors instead of opaque predictors.
Popular post-hoc explainers include local interpretable model-agnostic explanations (LIME)~\parencite{ribeiro2016should}, kernel Shapley additive explanations (KernelSHAP)~\parencite{Lundberg2017-ii}, and gradient-weighted class activation mapping (Grad-CAM)~\parencite{Selvaraju2020-fx}.
One advantage of using such post-hoc explainers is that they can focus on developing predictors that achieve the highest accuracy because they place few constraints on the architecture of the predictors. 
Inherently interpretable predictors make predictions and offer explanations for those predictions in a single model.
These include classical models, such as generalized additive models~\parencite{10.2307/2245459} and recent DNN-based models~\parencite{alvarez2018towards}. 

Explanations for predictions can be obtained using the above approaches.
Are these explanations appropriate?
There are several evaluation metrics that assess, from different perspectives, the quality of explanations. 
How correctly explanations reflect the predictor's behavior is measured with {\it faithfulness} of the explanations.
To evaluate the faithfulness, {\it insertion} and {\it deletion metrics} have been widely used in the literature~\parencite{petsiuk2018rise,Gevaert2022-tb}.
Intuitively, these metrics are calculated on the assumption that if the features, e.g., the pixels in an image, which are deemed important to the explanation are truly important to the predictor, the presence or absence of the features should strongly affect the output of the predictor.
If the insertion and deletion scores are both excellent, then we assume the explanation is faithful to the predictor.

The present study enables explainers to generate more faithful explanations with better insertion and deletion scores. 
To this end, we propose insertion/deletion metric-aware explanation-based optimization (ID-ExpO), which is a framework for optimizing predictors to improve both the insertion and deletion scores of the explanations produced by the explainers while maintaining the predictive accuracy of the predictors.
Because the original insertion and deletion metrics are non-differentiable with respect to the explanations and are directly unavailable for gradient-based optimization, we extend the metrics so that they are differentiable, and we use them to formalize the insertion and deletion metric-based regularizers.
By optimizing the predictors based on the standard prediction loss together with our regularizers simultaneously, ID-ExpO equips the predictors with capabilities that both produce accurate predictions and enable the explainers to produce more faithful explanations.
ID-ExpO can be applied to both post-hoc explainers and inherently interpretable models because it does not require any change in the architecture of the predictors.
In general, the post-hoc explainers are modeled differently than predictors.
For example, the LIME explainer approximates the predictor's behaviors using linear models around individual input data points, and the Grad-CAM explainer produces the explanation (saliency map) by aggregating the feature maps of the predictor's internal layer differently from the inference process of the predictor.
Owing to these differences, the explanations by the post-hoc explainers are likely not to reflect the actual feature contributions in the predictor.
Therefore, in this study, we focus on employing ID-ExpO to improve the post-hoc explainers and present its implementations for perturbation-based explainers, such as LIME and KernelSHAP, and gradient-based explainers, such as Grad-CAM.

In experiments on image and tabular datasets, we demonstrate the effectiveness of fine-tuning DNN predictors based on ID-ExpO compared with that of the existing stability-aware and fidelity-aware explanation-based optimization~\parencite{Plumb2019-mu} and the standard fine-tuning approach.
The experimental results show that ID-ExpO significantly improves the insertion and deletion scores on all the datasets while maintaining high predictive accuracy.
In a qualitative evaluation, we show that, by calculating our regularizers with only the top 30\% or 50\% of important features in explanations, the explanations for the predictor trained using ID-ExpO highlight the appropriate parts of features well.

\section{Related Work}\label{sec:RW}
\myparagraph{Existing Explainers.}
To interpret feature contributions in the individual predictions of complex machine learning models, various types of post-hoc explainers have been proposed, such as gradient-based explainers (including some of the CAM-based ones)~\parencite{Selvaraju2020-fx,chattopadhay2018grad,fu2020axiom,jiang2021layercam}, perturbation-based explainers~\parencite{ribeiro2016should,Lundberg2017-ii,zhao2021baylime}, and occlusion-based explainers~\parencite{petsiuk2018rise,wang2020score}.
One advantage of using the proposed method is that it enables an improvement in the explanations by the existing post-hoc explainers without changing their formulation, as long as the explainers are differentiable.
Some studies on the post-hoc explainers have proposed approaches that optimize or select explanations so that the features that are deemed important to the explanations contribute to better prediction~\parencite{petsiuk2018rise,fong2019understanding,Zhang2023-pq}.
However, \textcite{Zhang2023-pq} reported that their proposed method, one of those explainers, did not improve the insertion and deletion scores.
Another approach for this purpose is to use inherently interpretable predictors~\parencite{molnar2022}, including generalized additive models~\parencite{10.2307/2245459} and DNN-based models~\parencite{alvarez2018towards,agarwal2021neural,Yoshikawa2022-pu}, which can achieve both high predictive accuracy and transparency.
Several studies have proposed inherently interpretable DNN-based predictors whose attention maps and feature weights, which produce explanations, affect predictions and optimize them to improve the predictive accuracy~\parencite{Schwab2018GrangerCausalAM,fukui2019attention,Iida_2022_ACCV}.
Because explanations that lead to higher accuracy do not always result in better insertion and deletion scores, it is expected that the proposed method will also be helpful for the inherently interpretable predictors.

\myparagraph{Evaluation Metrics for Explanation.}
The ground truths of explanations are rarely observed because they are inside the complex predictor we would like to understand. %
Therefore, many studies have assessed explanations quantitatively using various proxy evaluation metrics~\parencite{Zhou2021-ac}.
In computer vision literature, insertion and deletion metrics are widely used to evaluate the faithfulness of the explanations~\parencite{petsiuk2018rise,Gevaert2022-tb}.
Several evaluation metrics that are related to insertion and deletion metrics have been proposed, such as sensitivity-$n$~\parencite{Ancona2018-di}, increase and drop rates~\parencite{chattopadhay2018grad,ramaswamy2020ablation}, and the iterative removal of features (IROF)~\parencite{rieger2020irof}.
For tabular data, stability~\parencite{alvarez2018towards}, sensitivity~\parencite{ghorbani2019interpretation}, and faithfulness~\parencite{bhatt2021evaluating}, which is another formulation for the insertion and deletion metrics, have been measured.
Although insertion and deletion metrics have not been used frequently for tabular data, employing these metrics on tabular data is beneficial for evaluating the combinatorial effects of features on prediction.

\myparagraph{Explanation-Based Optimization.}
Our study was inspired by the work of \textcite{Plumb2019-mu}.
They proposed an explanation-based optimization to improve the fidelity~\parencite{ribeiro2016should} and stability~\parencite{alvarez2018towards} of the explanations produced by perturbation-based post-hoc explainers for tabular data.
The main differences between the proposed method and their method are threefold:
The proposed method 1) aims to improve the faithfulness of the explanations by optimizing the insertion and deletion metrics, which are different from the fidelity and stability metrics; 2) is applicable to several data types, including images and tabular data; and 3) is effective for both perturbation-based and gradient-based explainers.
\textcite{Ismail2021-pk} proposed saliency-guided training, which optimized predictors such that the predictions between an original input and an input in which some of the pixels were masked according to the gradient-based explanations were similar.
Unlike our method, their method did not optimize the explanations to improve the insertion and deletion metrics.

\section{Proposed Method}\label{sec:proposed}
For the sake of concreteness, we consider a multiclass image classification task because the insertion and deletion metrics for this kind of task are widely used in the image domain.
Note that the proposed method can be applied to other data, such as text and tabular data, with slight changes.

\myparagraph{Problem Formulation.}
We are given an image $\mat{x} \in \set{X} \subseteq \mathbb{R}^{C \times H \times W}$ of the number of channels $C$, height $H$, and width $W$, which can be classified to a class within the set of classes $\set{Y} = \{1,2,\cdots,L\}$.
In addition, we have a pretrained trained predictor $f_{\theta}: \set{X} \rightarrow [0,1]^L$, e.g., a deep neural network, which outputs the probabilities of the classes, where $\theta$ is the set of model parameters.
We assume that the outputs of $f_{\theta}$ are normalized by the softmax function.
Next, we want to explain the prediction $\hat{y} = \mathrm{argmax}_l f_{\theta}(\bx)_l$ produced by $f_{\theta}$ using the given post-hoc explainer $e$, e.g., LIME, KernelSHAP, and Grad-CAM.
Here, the post-hoc explainer $e$ outputs pixel-level contributions (an explanation) $\bphi^{\hat{y}} \in \mathbb{R}^{H \times W}$ for image $\bx$ and predicted label $\hat{y}$, i.e., $\bphi^{\hat{y}} = e(\bx, f_{\theta}; \hat{y})$, where a larger positive value within $\bphi^{\hat{y}}$ means that its corresponding pixel is of greater importance to label $\hat{y}$.
In addition, in LIME and KernelSHAP, a large negative value in $\bphi^{\hat{y}}$ indicates that its corresponding pixel is not associated with label $\hat{y}$.
When we do not need to specify the label, we denote the explanation as $\bphi \in \mathbb{R}^{H \times W}$.  

With the above setup, our goal is to optimize (i.e., fine-tune) predictor $f_{\theta}$ on labeled data to force post-hoc explainer $e$ to produce faithful explanations with the best insertion and deletion scores while maintaining the predictor's inherent predictive capability.
To achieve our goal, we assume that predictor $f_{\theta}$ and post-hoc explainer $e$ are differentiable.
This assumption is commonly used because the DNN predictors trained through backpropagation are differentiable, and most post-hoc explainers, including perturbation-based, gradient-based, CAM-based, and occlusion-based explainers, are also differentiable. 

\myparagraph{Preliminaries: Insertion and Deletion Metrics.}
The insertion and deletion metrics are widely used to assess the faithfulness of an explanation.
In particular, the insertion metric evaluates the increase in the predicted probability for a target label when pixels that are deemed important to the explanation are gradually added to a blank image.
Conversely, the deletion metric evaluates the decrease in the predicted probability for the target label when such important pixels are gradually deleted from the input image.
Therefore, if the insertion score is high and the deletion score is low, we can say the explanation is faithful to the predictor.
More formally, for image $\bx \in \set{X}$ and label $y \in \set{Y}$, the insertion and deletion metrics are defined as follows:
\begin{align}
\small
\label{eq:proposed:ins}
&\mathtt{Ins}_{S}(\bx,y,\bphi^y,\bb; f_{\theta}) = \frac{1}{S} \sum_{s=1}^{S} f_{\theta}(\alpha(\bx;\bphi^y,\bb,s))_y,
\\
&\alpha(\bx;\bphi^y,\bb,s)_{ijk} = 
\begin{cases}
x_{ijk}, & (j,k) \in \argtopk(\bphi^y) \\
b_{ijk}, & \mbox{otherwise}
\end{cases},
\label{eq:proposed:ins:mask}
\\
\small
\label{eq:proposed:del}
&\mathtt{Del}_{S}(\bx,y,\bphi^y,\bb; f_{\theta}) = \frac{1}{S} \sum_{s=1}^{S} f_{\theta}(\beta(\bx;\bphi^y,\bb, s))_y,
\\
&\beta(\bx;\bphi^y,\bb,s)_{ijk} = 
\begin{cases}
b_{ijk}, & (j,k) \in \argtopk(\bphi^y) \\
x_{ijk}, & \mbox{otherwise}
\end{cases},
\label{eq:proposed:del:mask}
\end{align}
where $S \in \{1,2,\cdots,HW \}$ is the number of pixels that are used to evaluate these metrics. 
Although $S$ is typically set to $HW$, in some studies, $S$ is a smaller value than $HW$, e.g., 3.6\%, 30\%, and 50\% of $HW$~\parencite{zhang2021group,huber2023explainability,zeng2022abs}, because some of the pixels in the image are often critical to correct classification.
Here, $s$ can be incremented by a positive integer that is larger than one to reduce the number of times that $f_{\theta}$ is applied.
$\argtopk(\bphi^y)$ outputs a set of $s$ pairs of coordinate indices with the top-$s$ values in $\bphi^y$.
$\bb \in \set{X}$ is background values, e.g., channel-wise mean values calculated over all the training images.
$\alpha: \set{X} \rightarrow \set{X}$ and $\beta: \set{X} \rightarrow \set{X}$ are mask functions that mask a part of the pixels according to $\bphi^y$ and replace them with background values $\bb$.
The difference between these metrics lies in the choice of the mask functions. 
The insertion metric masks the pixels other than $\argtopk(\bphi^y)$ using $\alpha$, whereas the deletion metric masks the pixels in $\argtopk(\bphi^y)$ using $\beta$.
The insertion and deletion scores range between 0 and 1, where higher insertion scores and lower deletion scores, respectively, are better.

\begin{figure*}[t!]
\centering
\begin{minipage}[t]{0.34\textwidth}
\includegraphics[width=1.0\textwidth,pagebox=artbox]{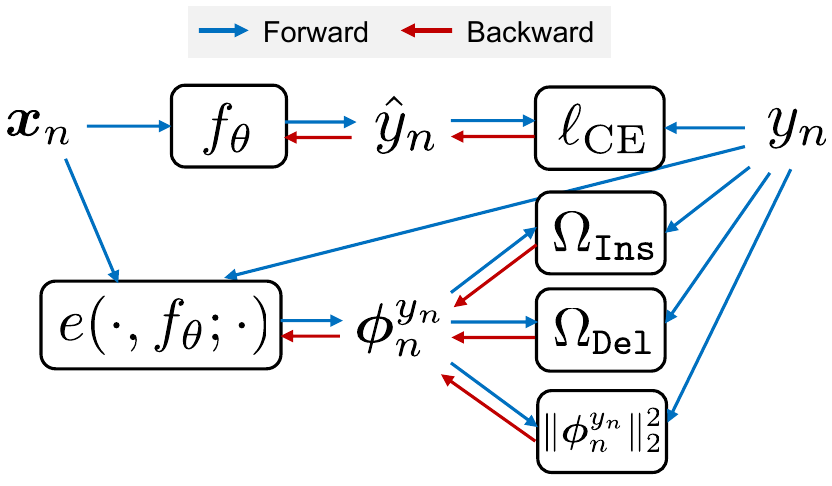}
\end{minipage}
\begin{minipage}[t]{0.34\textwidth}
\centering
\includegraphics[width=1.0\textwidth,pagebox=artbox]{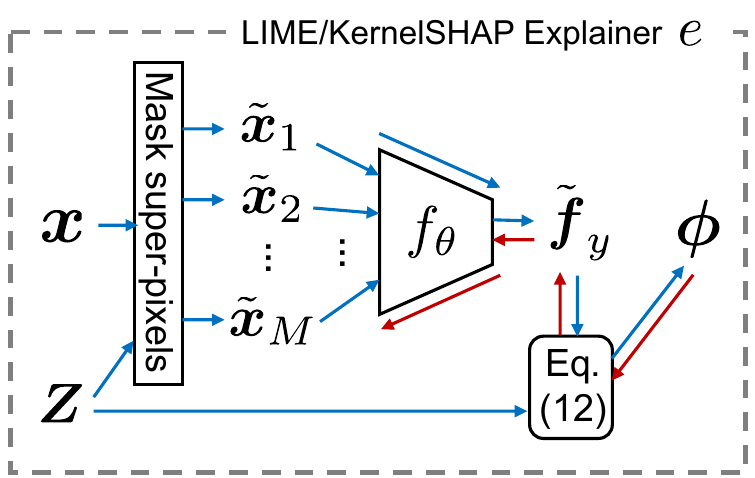}
\end{minipage}
\begin{minipage}[t]{0.29\textwidth}
\centering
\includegraphics[width=1.0\textwidth,pagebox=artbox]{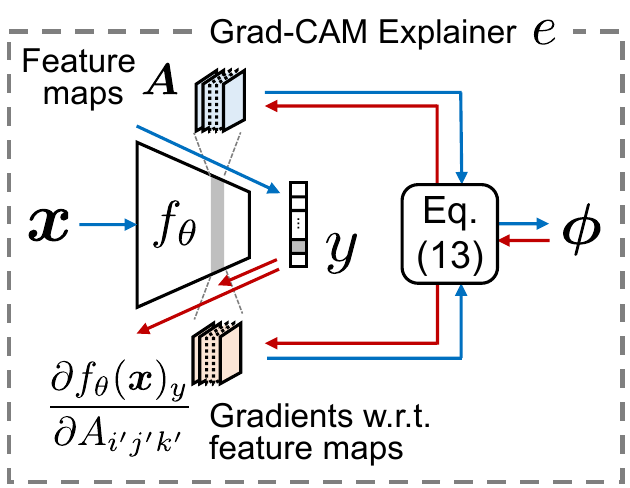}
\end{minipage}
\caption{
Overview of the forward and backward computations during training using ID-ExpO.
{\bf (Left)} the entire computational flow for each training sample $(\bx_n, y_n)$. 
The computation flows inside the {\bf (center)} LIME/KernelSHAP and {\bf (right)} Grad-CAM explainers.
Here, the red double line in Grad-CAM indicates that it computes second-order derivatives when it updates predictor $f_{\theta}$, as it uses the gradients w.r.t. feature maps to obtain $\bphi$.
}
\label{fig:proposed:forward_backward}
\end{figure*}

\subsection{Insertion/Deletion Metric-Aware Explanation-Based Optimization (ID-ExpO)}\label{sec:proposed:ID-ExpO}
Although the insertion and deletion metrics are used to assess explainers in the literature, we consider using them to optimize predictor $f_{\theta}$ such that the explanations by explainer $e$ have higher insertion scores and lower deletion scores.
We expect that such an optimization will furnish two benefits:
1) the explainer can produce feature contributions that are more faithful to the predictor's behaviors;
2) the explainer can clearly separate the important pixels from the less important ones in the explanation.
However, as these metrics are non-differentiable with respect to $\bphi$ due to the $\argtopk$ operation in the mask functions, we cannot use them directly to optimize $f_{\theta}$ with gradient-based optimization, such as stochastic gradient descent (SGD).

To solve this problem, we present differentiable insertion and deletion metrics with {\it soft} mask functions.
We rewrite the mask functions in~\bref{eq:proposed:ins:mask}~and~\bref{eq:proposed:del:mask} as follows:
\begin{align}
\small 
\alpha(\bx;\bphi,\bb,s)_{ijk} &= 
\begin{cases}
x_{ijk}, & \phi_{jk} \geq \topk(\bphi) \\
b_{ijk}, & \mbox{otherwise}
\end{cases},
\label{eq:proposed:mask1}
\\
\beta(\bx;\bphi,\bb,s)_{ijk} &= 
\begin{cases}
b_{ijk}, & \phi_{jk} \geq \topk(\bphi) \\
x_{ijk}, & \mbox{otherwise}
\end{cases},
\label{eq:proposed:mask2}
\end{align} %
where $\topk(\bphi)$ indicates the $s$th largest value in $\bphi$.
Here, the reformulation is equivalent to~\bref{eq:proposed:ins:mask}~and~\bref{eq:proposed:del:mask}, except in the case where the same value as $\topk(\bphi)$ exists in $\bphi$.
The mask functions~\bref{eq:proposed:mask1} and \bref{eq:proposed:mask2} are step functions that distinctly switch between $x_{ijk}$ and $b_{ijk}$ using the value of $\topk(\bphi)$ as a boundary, which are non-differentiable at the boundary and they have zero derivatives elsewhere.
To make them smooth, we approximate~\bref{eq:proposed:mask1} and \bref{eq:proposed:mask2} with soft step functions as follows:
\begin{align}
\label{eq:proposed:soft_alpha}
\alpha_{\mathrm{soft}}(\bx,\bphi;\bb,s)_{ijk} &= r(\phi_{jk};s) x_{ijk} + (1-r(\phi_{jk};s)) b_{ijk}, \\
\beta_{\mathrm{soft}}(\bx,\bphi;\bb,s)_{ijk} &= r(\phi_{jk};s) b_{ijk} + (1-r(\phi_{jk};s)) x_{ijk}, 
\label{eq:proposed:soft_beta}
\end{align}
where $r(\phi_{jk};s) \in [0, 1]$ is defined as $r(\phi_{jk};s)= \sigma\left(T \cdot (\phi_{jk} - t_s)\right)$.
$\sigma$ is a sigmoid function, $T > 0$ is a temperature parameter, and $t_s$ is the boundary value for the $s$-th largest value in $\bphi$ and is calculated as $t_s = \left(\topk(\bphi) + \topkplus(\bphi) \right) / 2$.
Here, if $T = \infty$, then \bref{eq:proposed:soft_alpha}~and~\bref{eq:proposed:soft_beta} are equivalent to~\bref{eq:proposed:mask1} and \bref{eq:proposed:mask2}, respectively.
Using~\bref{eq:proposed:soft_alpha}~and~\bref{eq:proposed:soft_beta} as the mask functions in~\bref{eq:proposed:ins}~and~\bref{eq:proposed:del}, respectively, we obtain the insertion and deletion metrics that are differentiable with respect to $\bphi$.

On the basis of the differentiable insertion and deletion metrics, we define insertion and deletion metric-based regularizers that regularize the predictor to maximize the insertion scores and minimize the deletion scores, as follows:
\begin{align}
\label{eq:proposed:ins_loss}
\Omega_{\Ins}(\bphi^y,f_{\theta};\bx,y,\bb) 
&= -\frac{1}{S}\sum_{s=1}^{S} \log f_{\theta}(\alpha_{\mathrm{soft}}(\bx,\bphi^y;\bb,s))_y, \\
\Omega_{\Del}(\bphi^y,f_{\theta};\bx,y,\bb) 
&= -\frac{1}{S}\sum_{s=1}^{S} \log \frac{f_{\theta}(\bx)_y}{f_{\theta}(\beta_{\mathrm{soft}}(\bx,\bphi^y;\bb,s))_y},
\label{eq:proposed:del_loss}
\end{align}
where we use $\log f_{\theta}$ instead of using $f_{\theta}$ directly for numerical stability.
We attempted three types of formulations using the deletion metric-based regularizer.
Consequently, \bref{eq:proposed:del_loss} achieved the high performance, as shown in Appendix~\appendixref{sec:appendix:del_loss}.

During the training, we used an $f_{\theta}$ that had been pretrained in a supervised learning manner, and we fine-tuned it using the regularizers together with the standard prediction loss.
In particular, given $N$ training samples $\{(\bx_n, y_n)\}_{n=1}^N$ where $\bx_n \in \set{X}$ and $y_n \in \set{Y}$, we solve the following minimization problem using SGD:
\begin{align}
\small
\argmin_{\theta} \sum_{n=1}^N \ell_{\mathrm{CE}}(f_{\theta}(\bx_n), y_n) 
+ \lambda_1 \Omega_{\Ins}(\bphi^{y_n}_n,f_{\theta};\bx_n,y_n,\bb) \nonumber\\
+ \lambda_2 \Omega_{\Del}(\bphi^{y_n}_n,f_{\theta};\bx_n,y_n,\bb)
+ \lambda_3 \|\bphi^{y_n}_n \|^2_2,
\label{eq:proposed:objective}
\end{align}
where $\ell_{\mathrm{CE}}(f_{\theta}(\bx_n), y_n)$ is the cross-entropy loss between the prediction $f_{\theta}(\bx_n)$ and the label $y_n$;
$\lambda_1,\lambda_2$, and $\lambda_3$ are hyperparameters;
$\lambda_1, \lambda_2 \geq 0$ are the weights for the regularizers, respectively. 
$\|\bphi^{y_n}_n \|^2_2$ is an L2 regularizer to prevent the divergence of $\bphi^{y_n}_n$, and $\lambda_3 \geq 0$ is its weight\footnote{Because our regularizers encourage separating important pixels from less important ones, the feature contributions for the important pixels may become larger, and those for the less important ones may become smaller (negatively larger).
Therefore, we add $\lambda_3 \|\bphi^{y_n}_n \|^2_2$ to avoid the divergence of $\bphi^{y_n}_n$, which may adversely affect the predictor's parameter update.}.
The left side of Figure~\ref{fig:proposed:forward_backward} illustrates an overview of forward and backward computations involved in executing~\bref{eq:proposed:objective}.

Because $\bphi^{y_n}_n$ is obtained using the post-hoc explainer $e(\bx_n, f_{\theta}; y_n)$, the implementation of~\bref{eq:proposed:objective} depends on which post-hoc explainer we use.
Below, we describe the implementations of perturbation-based and gradient-based explainers.

\myparagraph{ID-ExpO for Perturbation-Based Explainers.}
The perturbation-based explainers for an image learn interpretable functions that capture the relationship between the predictor's inputs and outputs using perturbations around the image.
The representative methods for the perturbation-based explanation are LIME and KernelSHAP, which calculate pixel-level contributions using the coefficients of linear regression models that have been learned on the perturbations around the input image.
We illustrate the computational flow of these explainers at the center of Figure~\ref{fig:proposed:forward_backward}.
First, we partition the image into $D$ superpixels.
Then, we generate $M$ binary random vectors where the $m$th vector is denoted by $\bz_m \in \{0,1\}^{D}$, and its $l$th element of that vector indicates whether its corresponding superpixel is masked ($z_{ml} = 0$) or not ($z_{ml} = 1$).
According to $\bz_m$, we obtain the masked perturbed image $\tilde{\bx}_m$ from the original image $\bx$.
Where $\bZ = [\bz_1,\bz_2,\cdots,\bz_M]^{\top} \in \{0,1\}^{M \times D}$ and $\tilde{\boldsymbol{f}}_{y} = [f_{\theta}(\tilde{\bx}_m)_y]_{m=1}^M$, the LIME and KernelSHAP explainers calculate the pixel-level contributions by first solving the weighted least-squares problem and then expanding the obtained coefficients into the pixels of the image, as follows:
\begin{equation}
e(\bx, f_{\theta}; y) = \expand_{D \rightarrow H \times W}\Big((\bZ^{\top} \bK \bZ + \epsilon \bI_D)^{-1} \bZ \bK \tilde{\boldsymbol{f}}_{y} \Big),
\label{eq:proposed:g_lime}
\end{equation}
where $\expand_{D \rightarrow H \times W}$ is a function that expands the contribution assigned to each of $D$ superpixels to the pixels associated with the superpixel.
$\bK$ is a diagonal matrix of size $M \times M$ whose $(m,m)$-element is the kernel value between a $D$-dimensional all-ones vector and $\bz_m$.
$\bI_D$ is an identity matrix of size $D$, and $\epsilon \geq 0$ is the hyperparameter of the L2 regularizer.
The key difference between LIME and KernelSHAP is the kernel function used to compute $\bK$: LIME uses an L2 kernel (for images), whereas KernelSHAP uses a Shapley kernel.

\myparagraph{ID-ExpO for Gradient-Based Explainers.}
One of the most popular gradient-based explainers is Grad-CAM.
We illustrate the computational flow of the Grad-CAM explainer on the right side of Figure~\ref{fig:proposed:forward_backward}.
Grad-CAM obtains nonnegative pixel-level contributions by calculating the activation map for the intermediate layers of predictor $f_{\theta}$.
Typically, convolutional neural networks are used as the predictors.
To compute the activation map, Grad-CAM uses the feature maps from a convolution layer of the CNN predictor, which are denoted by $\bA \in \mathbb{R}^{C' \times H' \times W'}$ where $C'$, $H'$ and $W'$ are the number of channels, height, and width of the feature maps, respectively.
In addition, it calculates the gradient for the output of the CNN predictor with respect to each element in the feature maps, i.e., $\frac{\partial f_{\theta}(\bx)_y}{\partial A_{i'j'k'}}$.
Using them, the Grad-CAM explainer calculates the activation map $\boldsymbol{M} \in \mathbb{R}_{\geq 0}^{H' \times W'}$ for label $y$, followed by the pixel-level importance from $\boldsymbol{M}$, as follows:
\begin{equation}
\label{eq:proposed:g_gradcam}
\small
e(\bx, f_{\theta}; y) 
= \expand_{H' \times W' \rightarrow H \times W}\Bigg(\underbrace{\mathrm{ReLU}\Big(\sum_{i'=1}^{C'} \omega^{y}_{i'} \bA_{i'} \Big)}_{= \boldsymbol{M}} \Bigg), 
\end{equation}
where
\begin{equation}
\omega^{y}_{i'} = \frac{1}{H' W'} \sum_{j'=1}^{H'} \sum_{k'=1}^{W'} \frac{\partial f_{\theta}(\bx)_y}{\partial A_{i'j'k'}}.
\end{equation}
Here, $\expand_{H' \times W' \rightarrow H \times W}$, as with that in~\bref{eq:proposed:g_lime}, expands the $(j',k')$-element of the activation map $\boldsymbol{M}$ to its corresponding pixels on the image; $\mathrm{ReLU}$ is the rectified linear unit.

\myparagraph{For Data Other Than Images.}
ID-ExpO is also available for cases where the input data is a $Q$-dimensional vector $\bx \in \set{X} \subseteq \mathbb{R}^Q$, e.g., text classification and tabular classification, whose $q$th dimension represents the $q$th feature value.
In these cases, we replace $ijk$ and $jk$ in \bref{eq:proposed:ins}--\bref{eq:proposed:del_loss} specifying indices in an image with $q \in \{1, 2, \cdots, Q\}$.
In addition, the maximum value of $S$ becomes $Q$ instead of $HW$.

\myparagraph{Missingness Bias.}
When we produce explanations by perturbation-based explainers and evaluate explanations using insertion/deletion metrics, we mask some of the pixels in the input image.
This input masking causes a problem called {\it missingness bias}, which means that the masked images are left out of the training input distribution.
\textcite{Jain2022-oe} reported that missingness bias causes the explanations by LIME not to be aligned with human intuition, and that training the model with random masking augmentation mitigates the problem.
Our regularizers,~\bref{eq:proposed:ins_loss}~and~\bref{eq:proposed:del_loss}, evaluate the predictive loss, $\log f(\cdot)$, when some of the input pixels are masked; this has effects similar to the training with random masking augmentation.
Therefore, our regularizers naturally manage the missingness bias.

\myparagraph{Computational Complexity.}
The computational time complexity of the forward computation for updating the predictor's parameters once is $O(B(E + S))$ in our framework, where $B$ is the batch size, $E$ is the computational time complexity for the generation of an explanation by the explainer, and $S$ is the number of times that the predictor is applied in Eqs.~\bref{eq:proposed:ins_loss} and \bref{eq:proposed:del_loss}.
$M$ depends on the explainer used.
For LIME and KernelSHAP, $E$ indicates the computational time complexity of Eq.~\bref{eq:proposed:g_lime}, $O(M + D^3)$, where $M$ is the number of predictions executed for perturbed images, and $D$ is the number of superpixels.
For Grad-CAM, $E$ indicates the computational time complexity of Eq.~\bref{eq:proposed:g_gradcam}, which is determined by the choice of predictor.

\section{Experiments}
We conducted experiments on two image datasets and six tabular datasets to evaluate the effectiveness of our ID-ExpO in the case of using LIME, KernelSHAP and Grad-CAM as post-hoc explainers.
Due to length limitations, we report the results of the image classification task in this section and those of the tabular classification task in Appendix~\appendixref{sec:appendix:tabular}.
Our implementation is based on PyTorch v.1.13, and the experiments were performed on a computer consisting of an Intel Xeon Platinum 8360Y CPU, an NVIDIA A100 SMX4 GPU, and 512 GB of RAM.

\myparagraph{Comparing Methods.}
We compared the proposed method (ID-ExpO) with the following methods: stability-aware explanation-based optimization (ExpO-S), fidelity-aware explanation-based optimization (ExpO-F), and fine-tuning without explanation regularizers ($\ell_{\mathrm{CE}}$-only).
ExpO-S and ExpO-F were based on the study in~\parencite{Plumb2019-mu}. 
They aimed to improve the stability and fidelity of explanations produced by perturbation-based explainers such as LIME.
ExpO-S learns the predictor with a regularizer so that the outputs of the predictor do not change for neighborhoods of the input.
ExpO-F learns the predictor with another regularizer so that the outputs of the predictor for the neighborhoods are fitted with a local linear function.
Because the original ExpO-S and ExpO-F were not intended for use on image data, we modified the ExpO-S and ExpO-F to apply to both image and tabular data, as described in Appendix~\appendixref{sec:appendix:image:comparing_method}. 
$\ell_{\mathrm{CE}}$-only learns the predictor without any regularizer of explanation, which is equivalent to optimizing~\bref{eq:proposed:objective} with $\lambda_1=\lambda_2=\lambda_3=0$.

\myparagraph{Evaluation.}
We quantitatively assessed the proposed method and the comparing methods in terms of predictive accuracy, insertion score~\bref{eq:proposed:ins}, and deletion score~\bref{eq:proposed:del}.
Here, we set $S$ to 30\% or 50\% of the number of features (pixels), which is the same value as that used by our regularizers~\bref{eq:proposed:ins_loss}~and~\bref{eq:proposed:del_loss}.
Instead of the deletion score, we use the one-minus-deletion score, which is calculated by subtracting the deletion score from one, for readability.
Furthermore, to check if the proposed method improves other faithfulness metrics that are not optimized directly, we also assessed explanations in sensitivity-$n$~\parencite{Ancona2018-di}, which evaluates how much, when $n$ features are randomly removed, the sum of the contributions of the removed features correlates with the decrease in the predicted confidence.

As the criterion used for model selection and for monitoring the progress of the training, we used the validation score function defined as follows:
\begin{equation}
\mathrm{valscore}(f_{\theta};\eta) = \eta\cdot \mathtt{Acc}(f_{\theta}) + \mathtt{Ins}(f_{\theta}) + 1 - \mathtt{Del}(f_{\theta}),
\label{eq:experiment:valscore}
\end{equation}
where $\mathtt{Acc}(f_{\theta})$, $\mathtt{Ins}(f_{\theta})$, and $\mathtt{Del}(f_{\theta})$ indicate predictive accuracy, average insertion score, and average deletion score for the validation set for the predictor with current parameters $f_{\theta}$, respectively.
$\eta \geq 0$ is an accuracy weight used to control the ratio of predictive and explanatory capabilities.
Unless otherwise noted, we set $\eta = 2$, which indicates that the predictive and explanatory capabilities should be equally evaluated.

\subsection{Image Classification}\label{sec:experiment:image}

\myparagraph{Datasets.}
We used two standard benchmark image classification datasets, CIFAR-10~\parencite{krizhevsky2009learning} and STL-10~\parencite{pmlr-v15-coates11a}.
CIFAR-10 contains 60,000 color images having a resolution of 32x32, classified into ten distinct classes.
There were originally 50,000 labeled images for training and 10,000 labeled images for testing.
In our experiment, we left 10,000 images in the original training set for validation.
STL-10 consists of color images having a resolution of 96x96, with ten classes and 1,300 images per class. 
It originally provided 5,000 labeled images for training and 8,000 labeled images for testing.
In our experiment, we left 500 images in the original training set for validation.

\myparagraph{Implementation.}
We used ResNet-18~\parencite{he2016deep} as a predictor.
We trained it in advance on each training set using the standard supervised learning, with data augmentation via random horizontal flipping and random cropping.
For the LIME explainer, we first partitioned the image into square-shaped superpixels of different sizes, depending on the dataset.
We used the superpixels of size 4x4 for 32x32 pixel images in CIFAR-10, and we used the superpixels of size 12x12 for the 96x96 pixel images in STL-10.
Therefore, the number of superpixels was set to $D=(32/4)^2=64$ for CIFAR-10 and $D=(96/12)^2=64$ for STL-10.
We also used $D$ as the constant to increment $s$ because each pixel in a superpixel has the same contribution value.
We generated $M=200$ perturbations and we set $\epsilon=0.01$ in~\bref{eq:proposed:g_lime}.
For the Grad-CAM explainer, we used the feature maps of the \texttt{conv3\_x} and \texttt{conv4\_x} building blocks in ResNet-18 for CIFAR-10 and STL-10, respectively.
The sizes of the feature maps were 8x8 for CIFAR-10 and 6x6 for STL-10. 
Depending on these sizes, we decided $s$ to increment by 16 for CIFAR-10 and by 256 for STL-10. 

The hyperparameters in ID-ExpO were $\lambda_1$, $\lambda_2$ and $\lambda_3$ in~\bref{eq:proposed:objective}, and the temperature parameter $T$.
For the experiments in this study, we used the same value $\lambda_{12}$ for $\lambda_1$ and $\lambda_2$.
We determined the best hyperparameter values in the ranges of $\lambda_{12} \in \{0.1,0.01,0.001 \}$ and $\lambda_3 \in \{0.001,0 \}$ based on~\bref{eq:experiment:valscore}. 
$T$ in the soft step function $r$ ensures that its value does not approach zero or one. 
Because the appropriate value of $T$ is different for different samples, we determined it as $T = \left(\frac{1}{\#(\bphi) - 1} \left(\max(\bphi) - \min(\bphi) \right)\right)^{-1}$ for each sample, where $\#(\bphi)$ is the number of elements with nonduplicated values in $\bphi$.
The hyperparameter in each of ExpO-S and ExpO-F, which is the strength of its own regularizer, is determined as the best one in the range of $\{0.1,0.01,0.001 \}$ based on~\bref{eq:experiment:valscore}.
There are no hyperparameters specific to $\ell_{\mathrm{CE}}$-only.
For the optimizer, we used an SGD optimizer with a minibatch size of 128, a momentum factor of 0.9, a weight decay of 0.0005, and Nesterov momentum. 
Its learning rate was determined on the basis of~\bref{eq:experiment:valscore} within the range of $\{10^{-4}, 10^{-5}\}$.
The training continued for 50 epochs or the value of $\mathrm{valscore}(f_{\theta};2)$ did not increase for ten consecutive epochs.

\begin{figure}[t!]
\centering
\begin{minipage}{0.48\hsize}
\includegraphics[width=\columnwidth]{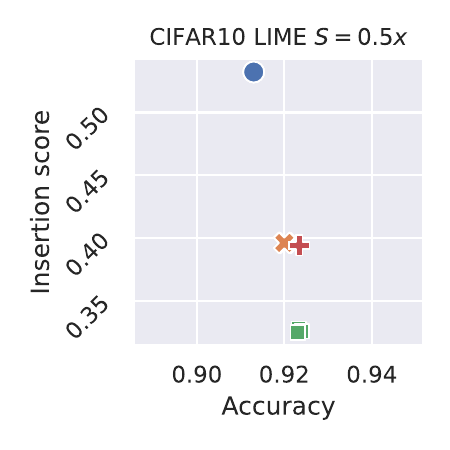}
\end{minipage}
\begin{minipage}{0.48\hsize}
\includegraphics[width=\columnwidth]{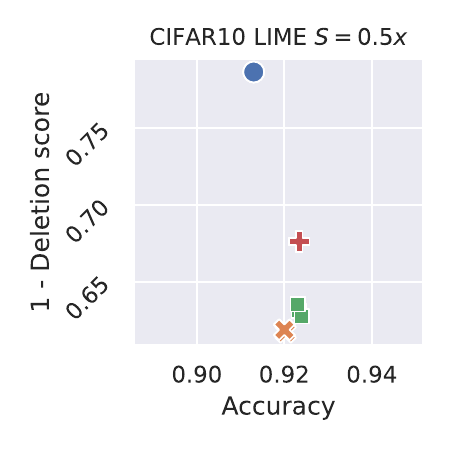}
\end{minipage}
\\
\begin{minipage}{0.48\hsize}
\includegraphics[width=\columnwidth]{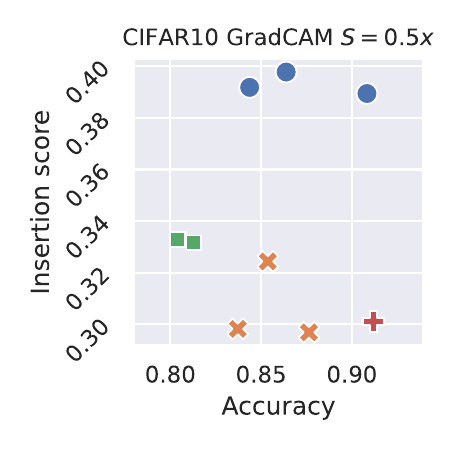} 
\end{minipage}
\begin{minipage}{0.48\hsize}
\includegraphics[width=\columnwidth]{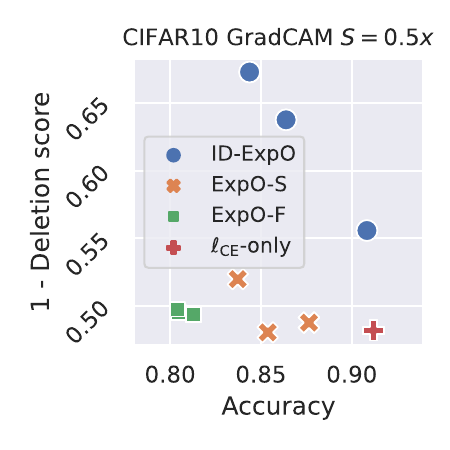} 
\end{minipage}
\caption{
  Mean insertion and mean one-minus-deletion scores against accuracy on CIFAR-10 in the case of $S = 0.5 \cdot HW$.
  The top row shows the results for LIME, and the bottom row shows the results for Grad-CAM.
  Each point indicates the result for the hyperparameters chosen on the basis of \bref{eq:experiment:valscore} with a different accuracy weight $\eta \in \{0.5, 1.0. \cdots, 3.0 \}$ (different $\eta$ values can be plotted in the same location).
  The higher the score, the better.
}
\label{fig:experiment:image_scores_0.5}
\end{figure}

\subsubsection{Results}
Figure~\ref{fig:experiment:image_scores_0.5} shows the insertion and one-minus-deletion scores against the predictive accuracy on CIFAR-10.
Here, the scores are averaged over all the samples in the test set.
ID-ExpO achieved the best insertion and deletion scores among the results.
This fact indicates that the insertion and deletion metric-based regularizers used in ID-ExpO are effective, although the stability-aware regularizer in ExpO-S and the fidelity-aware regularizer in ExpO-F are not suitable for improving those scores.
In terms of predictive accuracy, ID-ExpO was comparable with $\ell_{\mathrm{CE}}$-only when $\eta$ was controlled in~\bref{eq:experiment:valscore} to achieve the highest accuracy, e.g., for $\eta = 3$, the insertion and deletion scores of ID-ExpO were better than those of $\ell_{\mathrm{CE}}$-only.
Similar results were observed in the case of $S=0.3 \cdot HW$ and on STL-10, as shown in Appendix~\appendixref{sec:appendix:image:quantitative}.

As described in Section~\ref{sec:proposed:ID-ExpO}, the regularizers in ID-ExpO naturally make predictors robust to the missingness bias.
Missingness bias can unfairly make the insertion and deletion scores of the comparing methods better than those of ID-ExpO because the predictors that are sensitive to the bias may greatly change thier predictions, even if unimportant pixels are inserted in or masked out.
Therefore, the fact that ID-ExpO achieved the best insertion and deletion scores indicates that the regularizers in ID-ExpO are effective in improving the faithfulness of explanations.

\begin{figure}[t!]
\centering
\begin{minipage}{0.48\hsize}
\includegraphics[width=\columnwidth]{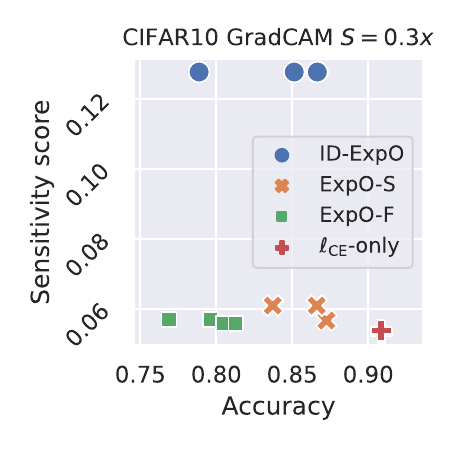}
\end{minipage}
\begin{minipage}{0.48\hsize}
\includegraphics[width=\columnwidth]{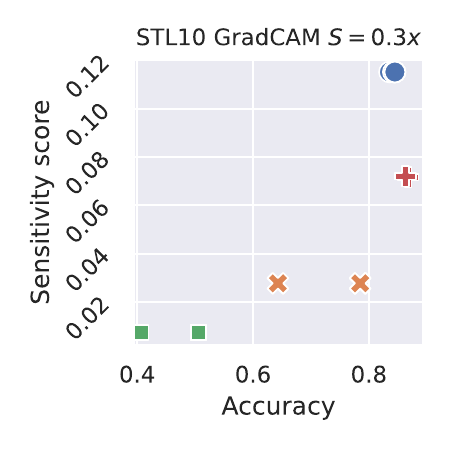}
\end{minipage}
\caption{
Mean sensitivity-$n$ scores against accuracy on CIFAR-10 (left) and STL-10 (right) in the case of $S = 0.3 \cdot HW$.
The higher the score, the better.
}
\label{fig:experiment:sensitivity_0.3}
\end{figure}

To investigate the impact of the proposed and the existing optimization methods on a faithfulness metric other than insertion/deletion metrics, we also evaluated the produced explanations in terms of sensitivity-$n$ metric in Figure~\ref{fig:experiment:sensitivity_0.3}, which ID-ExpO does not directly optimize\footnote{
Although the original sensitivity-$n$ evaluates feature contributions at a pixel level, it is not appropriate to assess the feature contributions produced by Grad-CAM and LIME, which are explained at a super-pixel level.
Therefore, we modified the sensitivity-$n$ so as to mask super-pixels randomly.
Here, we were set to $n=4$, i.e., we randomly masked four super-pixels instead of pixels in an image.}.
The figure shows that ID-ExpO consistently improved in terms of the sensitivity-$n$ metric on CIFAR-10 and STL-10, whereas ExpO-S and ExpO-F did not.
This result indicates that ID-ExpO does not overfit the insertion and deletion metrics, and can improve the faithfulness of explanations from various perspectives.

\begin{figure}[t!]
\begin{center}
\includegraphics[width=\columnwidth,pagebox=artbox]{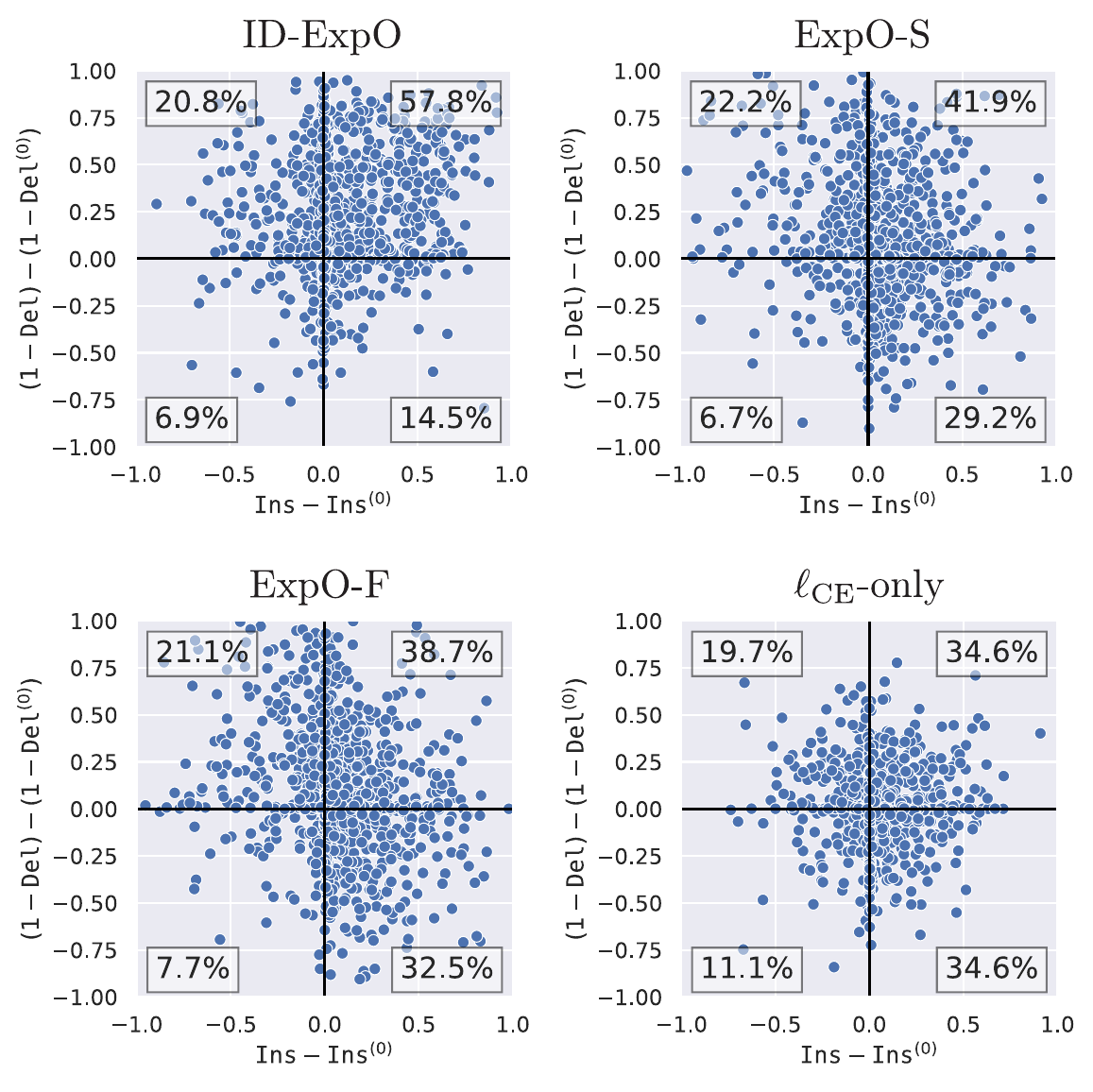}
\end{center}
\caption{
  Differences in the insertion and one-minus-deletion scores between before and after predictors were fine-tuned, with Grad-CAM using each method for 1,000 randomly selected individual test samples on CIFAR-10.
  $\Ins$ and $\Del$ indicate the mean insertion and deletion scores over the test set when the predictors are used after fine-tuning, whereas $\Ins^{(0)}$ and $\Del^{(0)}$ indicate the same scores before the fine-tuning.
  The percentage in each quadrant is the ratio of the samples located in the quadrant.
}
\label{fig:experiment:ins_vs_del}
\end{figure}

Figure~\ref{fig:experiment:ins_vs_del} shows how much the insertion and deletion scores of individual samples changed before and after we fine-tuned the predictor based on each method.
In ID-ExpO, ExpO-S, ExpO-F, and $\ell_{\mathrm{CE}}$-only, 57.8\%, 41.9\%, 38.7\% and 34.6\% of samples were located in the first quadrant, respectively.
Because the first quadrant means that the changes in the insertion and one-minus-deletion scores are both positive, we found that ID-ExpO was the most effective in making explanations more faithful.
Conversely, the ratios of the samples located in the third quadrant, which indicates that the samples became less faithful, were 6.9\%, 6.7\%, 7.7\% and 11.1\%, respectively.
We also found that the possibility that ID-ExpO would worsen the faithfulness of the explanations was relatively low.

\begin{figure*}[t!]
\centering
\includegraphics[width=0.85\textwidth]{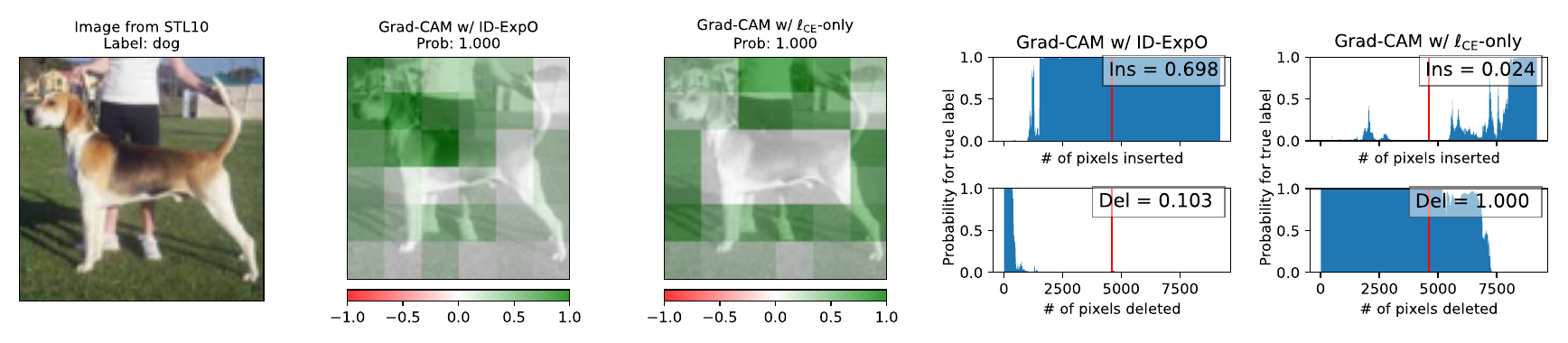} \\
\includegraphics[width=0.85\textwidth]{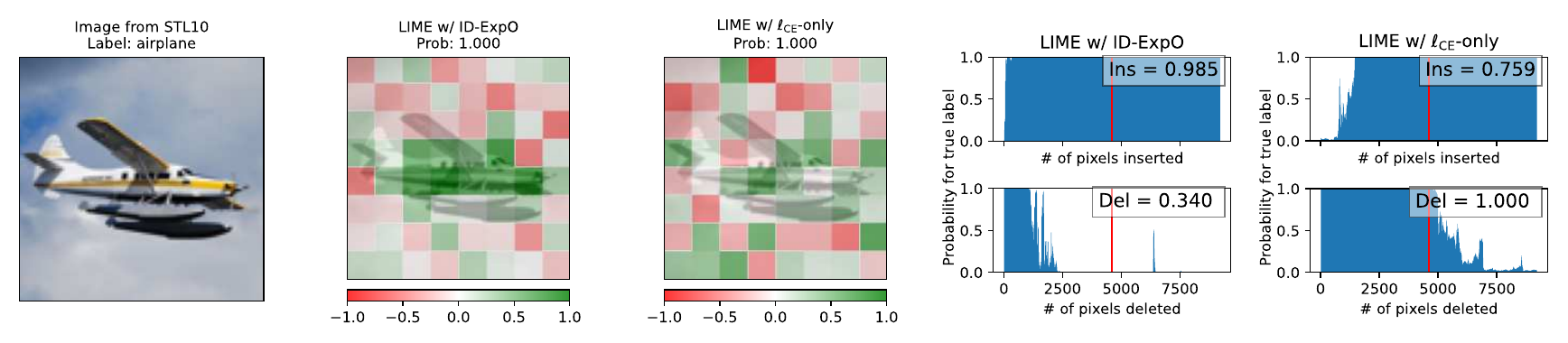} 
\\
{\small
\begin{minipage}{0.17\hsize}
\centering
  (a)
\end{minipage}
\begin{minipage}{0.17\hsize}
\centering
  (b)
\end{minipage}
\begin{minipage}{0.17\hsize}
\centering
  (c)
\end{minipage}
\begin{minipage}{0.17\hsize}
\centering
  (d)
\end{minipage}
\begin{minipage}{0.17\hsize}
\centering
  (e)
\end{minipage}
}
\caption{
Examples of the produced explanations for STL-10.
The first row shows the results obtained by Grad-CAM, while the other shows the results obtained by LIME.
Each row illustrates (a) an input image, (b)--(c) the heatmaps of the explanations by the explainers with ID-ExpO and $\ell_{\mathrm{CE}}$-only, and (d)--(e) the insertion score (top) and the deletion score (bottom) for those explanations in the case of $S=0.5\cdot HW$, which means that the scores are the blue areas to the left of red vertical lines.
}
\label{fig:experiment:image_examples}
\end{figure*}

In Figure~\ref{fig:experiment:image_examples}, we visualize the produced explanations for the predictors trained using ID-ExpO and $\ell_{\mathrm{CE}}$-only, and we show the distributions of the insertion and deletion metrics for the explanations.
More results are shown in Appendix~\appendixref{sec:appendix:image_examples}.
Here, we show only the examples in Figure~\ref{fig:experiment:image_examples}, in which correct prediction was made.
As shown in Figures~\ref{fig:experiment:image_examples}(d)~and~\ref{fig:experiment:image_examples}(e), the distributions of the insertion and deletion metrics were quite different between ID-ExpO and $\ell_{\mathrm{CE}}$-only.
This result indicates that our insertion and deletion metric-aware regularizers positively affected those distributions, as we expected.
In addition, as shown in Figures~\ref{fig:experiment:image_examples}(b)~and~\ref{fig:experiment:image_examples}(c), we found that the explanations differed from each other.
In particular, in the heatmap explanations for ID-ExpO (Figure~\ref{fig:experiment:image_examples}(b)), large positive contributions were assigned to some of the superpixels that captured the object of the class label well.
This is because the evaluation of the insertion and deletion scores in our regularizers is truncated at 30\% or 50\% of the number of pixels.
These results indicate that ID-ExpO can change the explanations to preferentially assign larger positive contributions to the pixels that strongly affect the prediction to improve those metrics.

\myparagraph{Comparison with Adversarially Robust Models.}
\textcite{Shah2021-mh} reported that adversarially robust models could make explanations based on input gradients more faithful to the model predictions than could models trained in the standard supervised learning manner.
However, it is not clear whether the adversarially robust models were helpful in producing faithful explanations when LIME and Grad-CAM were used.
To compare the models trained with ID-ExpO and with the adversarially robust models, we evaluated the adversarially robust ResNet-50 model (ADV for short) used in the work of \textcite{Shah2021-mh} in terms of insertion and one-minus-deletion metrics.
Here, the weights for ADV that are trained on CIFAR-10 are publicly available, and the predictive accuracy of ADV was 0.854 in our setting, which was comparable to or lower than that of the ResNet-18 model that had been fine-tuned with ID-ExpO.
The insertion and one-minus-deletion scores of ADV were 0.282 and 0.567 when Grad-CAM was used, and they were 0.456 and 0.778 when LIME was used.
Compared with the results shown in Figure~\ref{fig:experiment:image_scores_0.5}, we found that in terms of the insertion scores, the model trained with ID-ExpO significantly outperformed ADV, and in terms of the one-minus-deletion scores, the model trained with ID-ExpO was comparable to or better than ADV.
This result indicates that for Grad-CAM and LIME, ID-ExpO was more effective than the adversarially robust model in improving the faithfulness of explanations.

\section{Conclusion}
We proposed an explanation-based optimization method that learns machine learning predictors, such as DNNs, with insertion and deletion metrics-aware regularizers.
By fine-tuning the predictors based on the proposed method, we were able to confirm that several explainers, including perturbation-based and gradient-based explainers, could produce explanations that were faithful to the predictors' behaviors.
In future work, we will further verify the effectiveness of our insertion and deletion metrics-aware regularizers in improving the faithfulness of the explanations made by inherently interpretable models~\parencite{alvarez2018towards,Yoshikawa2021-ef} and parameterized explainers~\parencite{Situ2021-yo}.

\acknowledgments{
We thank the anonymous reviewers for their valuable suggestions. This work was supported by JSPS KAKENHI Grant Number 22K17953.}

\printbibliography%

\section*{Checklist}

 \begin{enumerate}

 \item For all models and algorithms presented, check if you include:
 \begin{enumerate}
   \item A clear description of the mathematical setting, assumptions, algorithm, and/or model. {\bf [Yes]}
   \item An analysis of the properties and complexity (time, space, sample size) of any algorithm. {\bf [Yes]}
   \item (Optional) Anonymized source code, with specification of all dependencies, including external libraries. {\bf [No]} However, we will release the source code after the paper is accepted.
 \end{enumerate}

 \item For any theoretical claim, check if you include:
 \begin{enumerate}
   \item Statements of the full set of assumptions of all theoretical results. {\bf [Not Applicable]}
   \item Complete proofs of all theoretical results. {\bf [Not Applicable]}
   \item Clear explanations of any assumptions. {\bf [Not Applicable]}
 \end{enumerate}

 \item For all figures and tables that present empirical results, check if you include:
 \begin{enumerate}
   \item The code, data, and instructions needed to reproduce the main experimental results (either in the supplemental material or as a URL). {\bf [Yes]}
   \item All the training details (e.g., data splits, hyperparameters, how they were chosen). {\bf [Yes]}
   \item A clear definition of the specific measure or statistics and error bars (e.g., with respect to the random seed after running experiments multiple times). {\bf [Yes]}
   \item A description of the computing infrastructure used. (e.g., type of GPUs, internal cluster, or cloud provider). {\bf [Yes]}
 \end{enumerate}

 \item If you are using existing assets (e.g., code, data, models) or curating/releasing new assets, check if you include:
 \begin{enumerate}
   \item Citations of the creator If your work uses existing assets. {\bf [Yes]}
   \item The license information of the assets, if applicable. {\bf [Not Applicable]}
   \item New assets either in the supplemental material or as a URL, if applicable. {\bf [Not Applicable]}
   \item Information about consent from data providers/curators. {\bf [Not Applicable]}
   \item Discussion of sensible content if applicable, e.g., personally identifiable information or offensive content. {\bf [Not Applicable]}
 \end{enumerate}

 \item If you used crowdsourcing or conducted research with human subjects, check if you include:
 \begin{enumerate}
   \item The full text of instructions given to participants and screenshots. {\bf [Not Applicable]}
   \item Descriptions of potential participant risks, with links to Institutional Review Board (IRB) approvals if applicable. {\bf [Not Applicable]}
   \item The estimated hourly wage paid to participants and the total amount spent on participant compensation. {\bf [Not Applicable]}
 \end{enumerate}

 \end{enumerate}

\pagebreak
\documentclass[twoside]{article}

\usepackage{aistats2024}

\usepackage[utf8]{inputenc} %
\usepackage[T1]{fontenc}    %
\usepackage{hyperref}       %
\usepackage{url}            %
\usepackage{booktabs}       %
\usepackage{amsfonts}       %
\usepackage{nicefrac}       %
\usepackage{microtype}      %
\usepackage{xcolor}         %
\usepackage{amsmath}
\usepackage{graphicx}
\usepackage{multirow}
\usepackage{wrapfig}
\usepackage[style=numeric,sorting=none]{biblatex}%
\usepackage{titlesec}
\usepackage{caption}

\newcommand{\myparagraph}[1]{{\bf #1}\ \ }
\newcommand{\appendixref}[1]{\ref*{#1} of the supplementary material}
\newcommand{\mainornot}[2]{\ifismaindoc#1\else#2\fi}

\newcommand{\argmax}{\mathop{\rm argmax}\limits}
\newcommand{\argmin}{\mathop{\rm argmin}\limits}
\newcommand{\argtopk}{\mbox{arg top-{\it s}}}
\newcommand{\topk}{\mbox{{\it s}th-val}}
\newcommand{\topkplus}{\mbox{({\it s}+1)th-val}}
\newcommand{\mat}[1]{\boldsymbol{#1}}
\newcommand{\bmc}[1]{\boldsymbol{\mathcal{#1}}}
\newcommand{\bref}[1]{(\ref{#1})}
\newcommand{\set}[1]{\mathcal{#1}}
\newcommand{\bA}{\boldsymbol{A}}
\newcommand{\bx}{\boldsymbol{x}}
\newcommand{\by}{\boldsymbol{y}}
\newcommand{\bz}{\boldsymbol{z}}
\newcommand{\bZ}{\boldsymbol{Z}}
\newcommand{\bK}{\boldsymbol{K}}
\newcommand{\bI}{\boldsymbol{I}}
\newcommand{\bb}{\boldsymbol{b}}
\newcommand{\br}{\boldsymbol{r}}
\newcommand{\bphi}{\boldsymbol{\phi}}
\newcommand{\Del}{\mathtt{Del}}
\newcommand{\Ins}{\mathtt{Ins}}
\newcommand{\CDel}{\mathtt{CDel}}
\newcommand{\CIns}{\mathtt{CIns}}
\newcommand{\dist}{\mathtt{dist}}
\newcommand{\expand}{\mathtt{expand}}

\addbibresource{idexpo_references.bib}%
\AtBeginBibliography{\small}

\newif\ifismaindoc
\ismaindocfalse %

\begin{document}

\onecolumn

\ifismaindoc
\else
  \aistatstitle{Supplementary Material:\\ Explanation-Based Training with Differentiable Insertion/Deletion Metric-Aware Regularizers}
\fi

\appendix

\setcounter{equation}{0}
\setcounter{figure}{0}
\setcounter{table}{0}
\renewcommand{\theequation}{A.\arabic{equation}}
\renewcommand{\thefigure}{A.\arabic{figure}}
\renewcommand{\thetable}{A.\arabic{table}}

\begin{figure}[h]
\begin{minipage}{0.245\hsize}
\includegraphics[width=\textwidth]{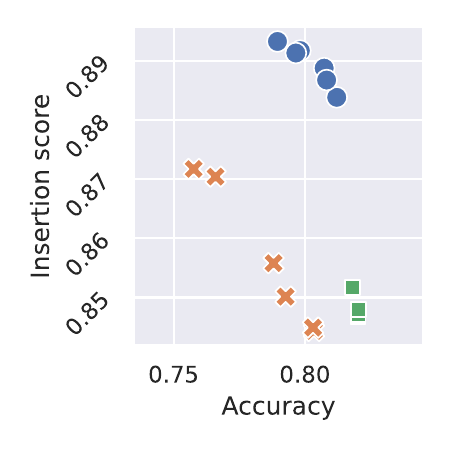}
\end{minipage}
\begin{minipage}{0.245\hsize}
\includegraphics[width=\textwidth]{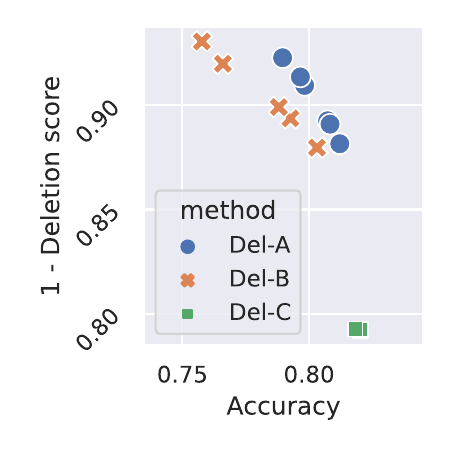}
\end{minipage}
\begin{minipage}{0.245\hsize}
\includegraphics[width=\textwidth]{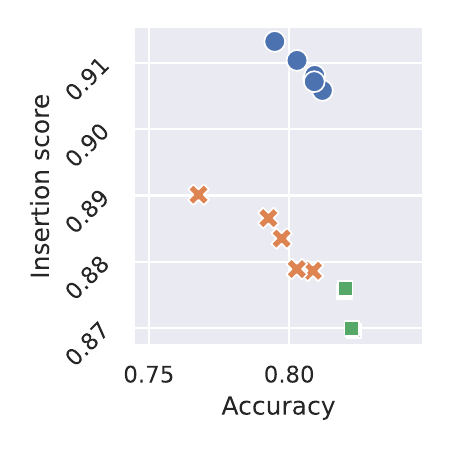} 
\end{minipage}
\begin{minipage}{0.245\hsize}
\includegraphics[width=\textwidth]{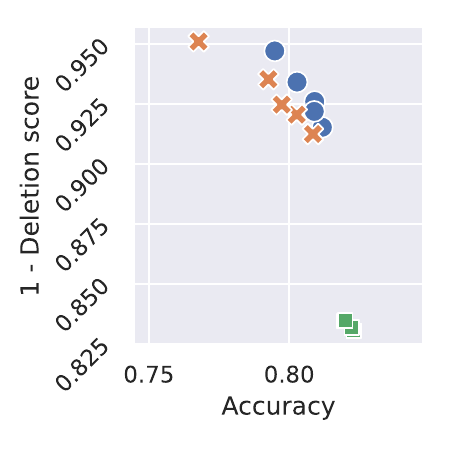} 
\end{minipage}
\caption{
  Mean insertion and mean one-minus-deletion scores against mean accuracy averaged over six tabular datasets when LIME was used as the explainer among three types of deletion metric-aware regularizers.
  Each point has a different accuracy weight $\eta \in \{0.5, 1.0. \cdots, 3.0 \}$.
  Figures of the first two columns are the results at $S=0.3 \cdot HW$, while those of the other columns are the results at $S=0.5 \cdot HW$.
  The higher the score, the better.
}
\label{fig:appendix:del_losses}
\end{figure}

\section{Comparison Among Different Types of Deletion Metric-Based Regularizers}\label{sec:appendix:del_loss}

We formalized three types of deletion metric-based regularizers, as follows:
\begin{align}
\mbox{\bf Del-A}:\quad 
\Omega_{\Del}(\bphi^y,f_{\theta};\bx,y,\bb) 
&= -\frac{1}{S}\sum_{s=1}^{S} \log \frac{f_{\theta}(\bx)_y}{f_{\theta}(\beta_{\mathrm{soft}}(\bx,\bphi^y;\bb,s))_y}, \\
\mbox{\bf Del-B}:\quad 
\Omega_{\Del}(\bphi^y,f_{\theta};\bx,y,\bb) 
&= \frac{1}{S}\sum_{s=1}^{S} \log f_{\theta}(\beta_{\mathrm{soft}}(\bx,\bphi^y;\bb,s))_y, \\
\mbox{\bf Del-C}:\quad 
\Omega_{\Del}(\bphi^y,f_{\theta};\bx,y,\bb) 
&= -\frac{1}{S}\sum_{s=1}^{S} \log \left(1 - f_{\theta}(\beta_{\mathrm{soft}}(\bx,\bphi^y;\bb,s))_y \right), 
\end{align}
where Del-A is what the proposed method employs.
Del-B is similar to Del-A, but does not have the term improving the prediction for the original input.
Del-C is similar to Del-B, but is formalized like the term for negative class in the binary cross entropy loss.

Figure~\ref{fig:appendix:del_losses} shows the insertion and one-minus-deletion scores of the three formulations.
We found that Del-A achieved the best balance of the accuracy and insertion/one-minus-deletion scores.

\section{On Experiments on Image Datasets}\label{sec:appendix:image}
\subsection{Modified Version of ExpO-S and ExpO-F}\label{sec:appendix:image:comparing_method}

The original ExpO-Stability (ExpO-S for short) and ExpO-Fidelity (ExpO-F for short) aim at making predictions and their corresponding explanations robust to slight changes in the feature values of the input, respectively~\parencite{Plumb2019-mu}.
The authors stated that the ExpO-F did not evaluate for non-semantic features, such as images, as the fidelity metric is not appropriate for the non-semantic features~\parencite[Appendix A.8]{Plumb2019-mu}.
Therefore, while keeping the idea of the original ExpO-F, we modified it to apply it to image data.
In particular, we utilize the same approach to LIME for image data, which we describe in Section~\mainornot{\ref{sec:proposed:ID-ExpO}}{3.1}, as the fidelity regularizer mimics the derivation of the explanations produced by LIME.
Below we use the same notation of the variables in Section~\mainornot{\ref{sec:proposed:ID-ExpO}}{3.1}.
First, for a given training sample $(\bx, y)$, we obtain a LIME explanation $\bphi^y$ by applying~\mainornot{\bref{eq:proposed:g_lime}}{(9)}.
Then, since $\bphi^y$ is the coefficients of a local linear model around input $\bx$, we can calculate the fidelity-aware regularizer based on the original ExpO-F as follows:
\begin{equation}
\Omega_{\mathtt{Fidelity}}(\bphi^y,f_{\theta};\bx,y,\{\bz_m\}_{m=1}^M,\bK) 
= \frac{1}{M} \sum_{m=1}^M \bK_{mm} (f_{\theta}(\tilde{\bx}_m)_y - \bphi^y{}^\top \bz_m)^2.
\label{eq:appendix:expo-f}
\end{equation}
where $\bz_m \in \{0,1\}^D$ is the $m$th binary random vector to mask $D$ super-pixels, $f_{\theta}(\tilde{\bx}_m)_y$ is the predicted probability of label $y$ for perturbed mask image $\tilde{\bx}_m$, $\bK_{mm}$ is the value of a cosine kernel between a $D$-dimensional all-one vector and $\bz_m$.

The original ExpO-S can be applied to numerical features and non-semantic features as it calculates the differences between the prediction for the original input and that for the perturbed input, in which Gaussian perturbations are added to the features of the input.
However, for the consistency of evaluation, we modified the original ExpO-S to perturb the input with binary random mask vectors $\bZ$.
In particular, we calculate the stability-aware regularizer as follows:
\begin{equation}
\Omega_{\mathtt{Stability}}(\bphi^y,f_{\theta};\bx,y,\{ \tilde{\bx}_m \}_{m=1}^M,\bK) 
= \frac{1}{M} \sum_{m=1}^M \bK_{mm} (f_{\theta}(\tilde{\bx}_m)_y - f_{\theta}(\bx))^2.
\end{equation}

\subsection{All Quantitative Results on Image Datasets}\label{sec:appendix:image:quantitative}

\begin{figure}[p!]
\small
{\bf (A) LIME on CIFAR-10}\\
\begin{minipage}{0.245\hsize}
\includegraphics[width=\textwidth]{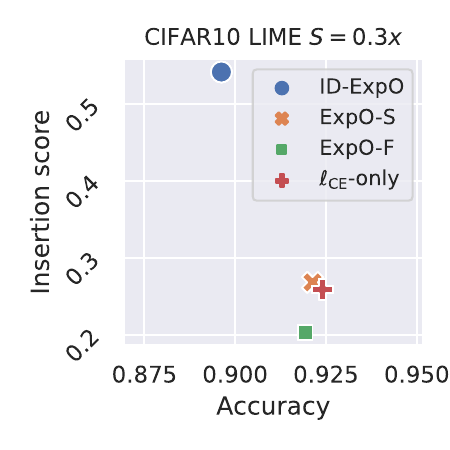}
\end{minipage}
\begin{minipage}{0.245\hsize}
\includegraphics[width=\textwidth]{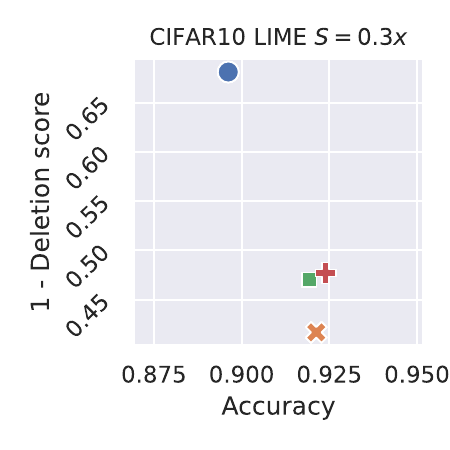}
\end{minipage}
\begin{minipage}{0.245\hsize}
\includegraphics[width=\textwidth]{fig/image_scatter_outlined/scatter_cifar10_lime_K0.5_acc_ins.pdf}
\end{minipage}
\begin{minipage}{0.245\hsize}
\includegraphics[width=\textwidth]{fig/image_scatter_outlined/scatter_cifar10_lime_K0.5_acc_del.pdf}
\end{minipage}
\\
{\bf (B) Grad-CAM on CIFAR-10}\\
\begin{minipage}{0.245\hsize}
\includegraphics[width=\textwidth]{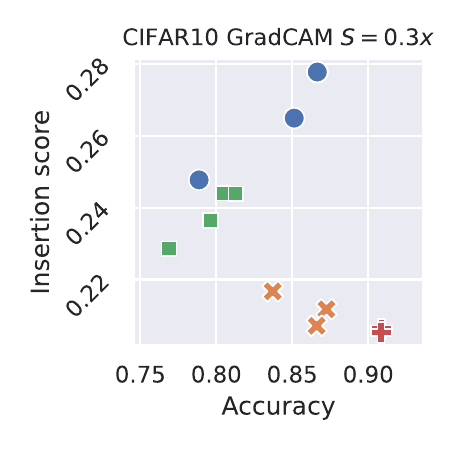}
\end{minipage}
\begin{minipage}{0.245\hsize}
\includegraphics[width=\textwidth]{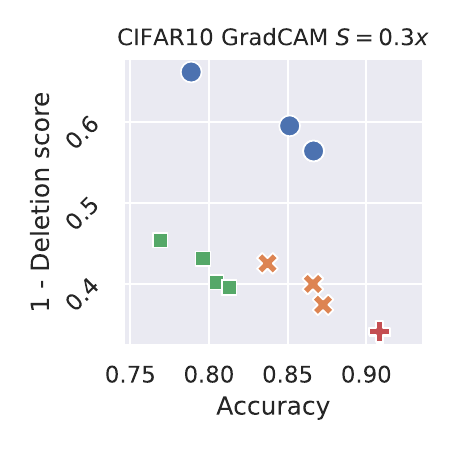}
\end{minipage}
\begin{minipage}{0.245\hsize}
\includegraphics[width=\textwidth]{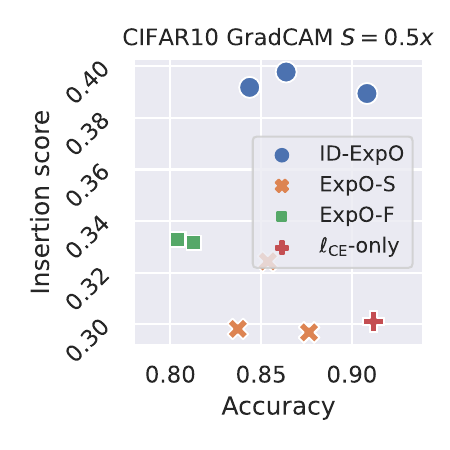}
\end{minipage}
\begin{minipage}{0.245\hsize}
\includegraphics[width=\textwidth]{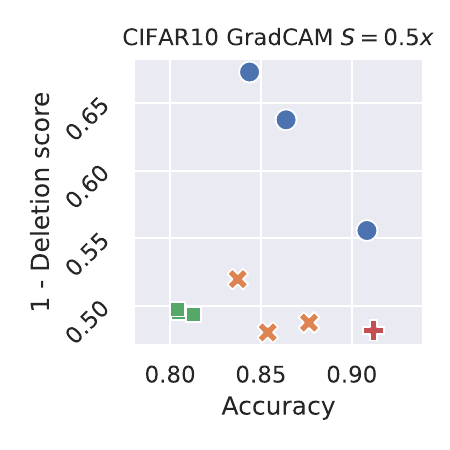}
\end{minipage}
\\
{\bf (C) LIME on STL-10}\\
\begin{minipage}{0.245\hsize}
\includegraphics[width=\textwidth]{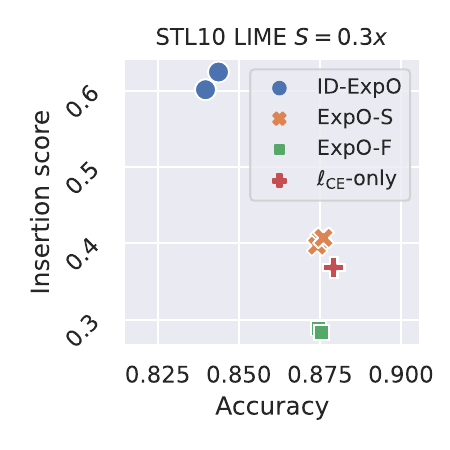}
\end{minipage}
\begin{minipage}{0.245\hsize}
\includegraphics[width=\textwidth]{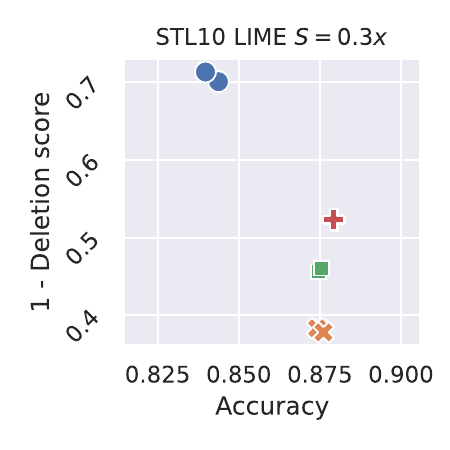}
\end{minipage}
\begin{minipage}{0.245\hsize}
\includegraphics[width=\textwidth]{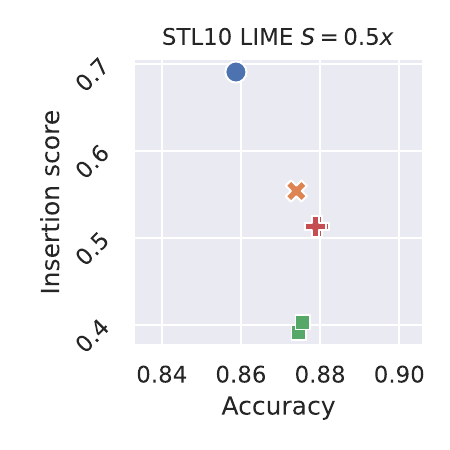}
\end{minipage}
\begin{minipage}{0.245\hsize}
\includegraphics[width=\textwidth]{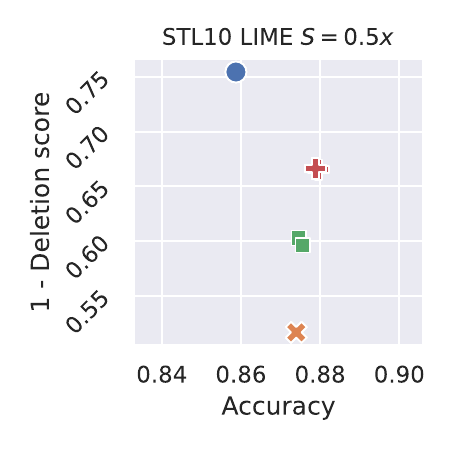}
\end{minipage}
\\
{\bf (D) Grad-CAM on STL-10}\\
\begin{minipage}{0.245\hsize}
\includegraphics[width=\textwidth]{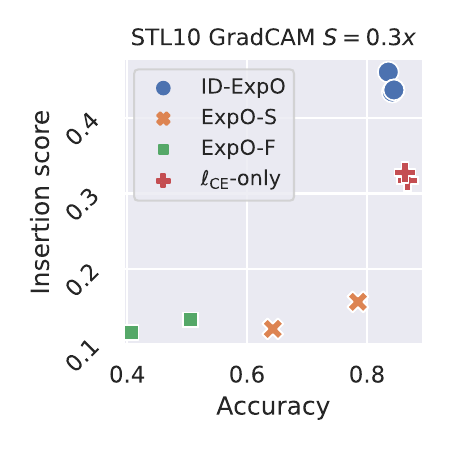}
\end{minipage}
\begin{minipage}{0.245\hsize}
\includegraphics[width=\textwidth]{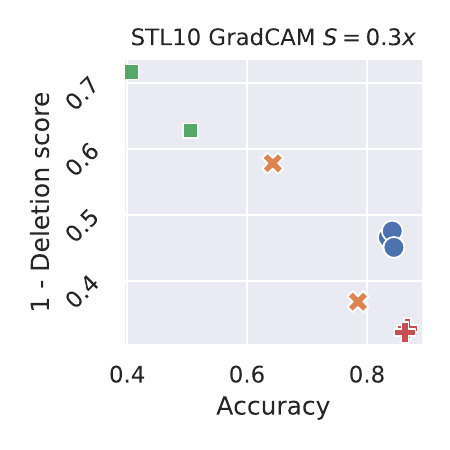}
\end{minipage}
\begin{minipage}{0.245\hsize}
\includegraphics[width=\textwidth]{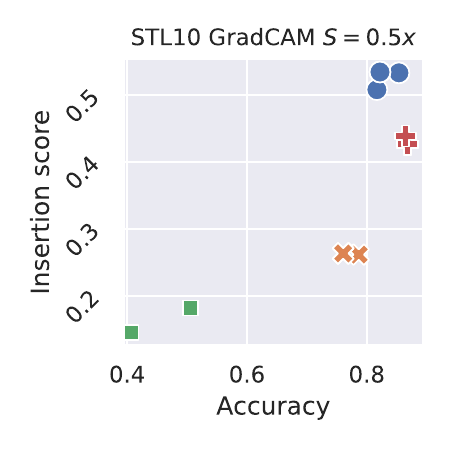}
\end{minipage}
\begin{minipage}{0.245\hsize}
\includegraphics[width=\textwidth]{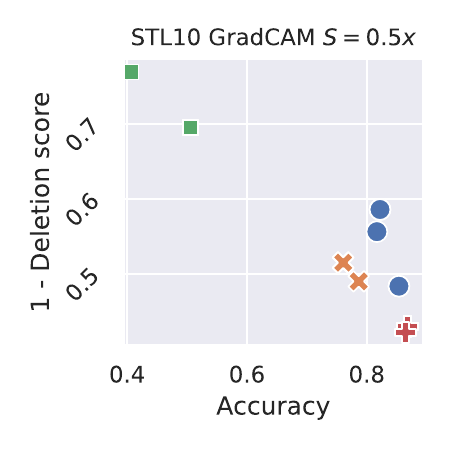}
\end{minipage}
\caption{
  Mean insertion and mean one-minus-deletion scores against accuracy on each image dataset.
  Each row indicates a different pair of an explainer and a dataset.
  The first two columns show the results when $S=0.3 \cdot HW$, while the last two columns show the results when $S=0.5 \cdot HW$.
  Each point indicates the result for the hyperparameters chosen based on \mainornot{\bref{eq:experiment:valscore}}{(11)} with a different accuracy weight $\eta \in \{0.5, 1.0. \cdots, 3.0 \}$ (different $\eta$ values can be plotted in the same location).
  The higher the score, the better.
}
\label{fig:appendix:image_scores}
\end{figure}

Figure~\ref{fig:appendix:image_scores} shows the insertion and the one-minus-deletion scores against the accuracy on each image dataset.
On all the explainers and the datasets except Grad-CAM on STL-10 (Figure~\ref{fig:appendix:image_scores}(D)), we found that the insertion and one-minus-deletion scores of ID-ExpO are superior to those of the other methods.
With the accuracy, by putting emphasis on the accuracy, i.e., by setting $\eta = 3$, we found that ID-ExpO in the case of $S=0.5\cdot HW$ could keep comparable or slightly low accuracy compared to the others, while the highest insertion and one-minus-deletion scores.
However, in the case of $S=0.3\cdot HW$, we found that the accuracy of ID-ExpO was lower than the other methods and the accuracy of ID-ExpO in the case of $S=0.5\cdot HW$.
The setting of $S=0.3\cdot HW$ means that our insertion and deletion metric-aware regularizers force the predictor to use only 30\% of the pixels in an image in prediction.
The result indicates that the regularization was too strong to predict correctly for the image datasets.

\subsection{Additional Examples of Produced Explanations on Image Datasets}\label{sec:appendix:image_examples}

\begin{figure}[p!]
\small
{\bf (A) LIME on CIFAR-10 ($S=0.5 \cdot HW$)}
\begin{center}
\includegraphics[width=0.85\textwidth]{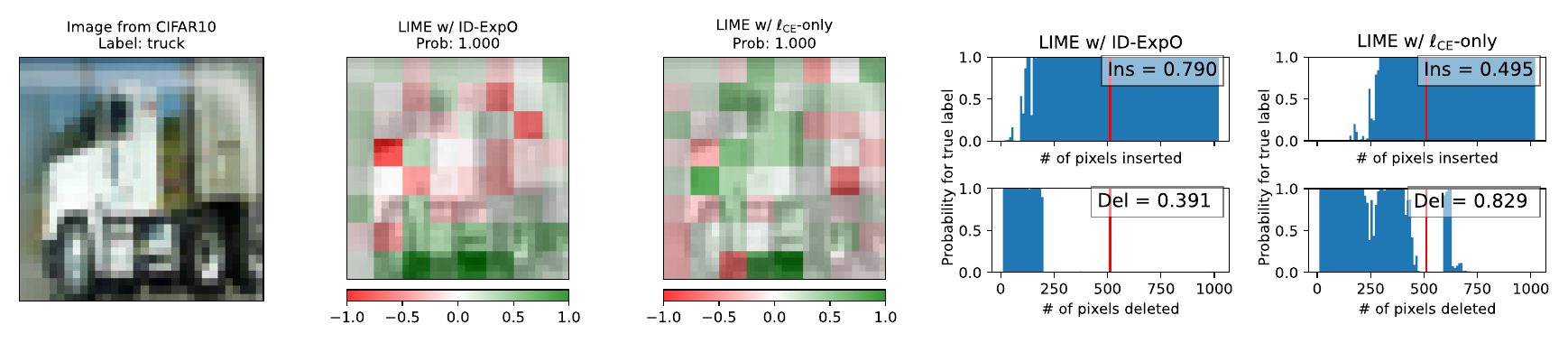} \\
\includegraphics[width=0.85\textwidth]{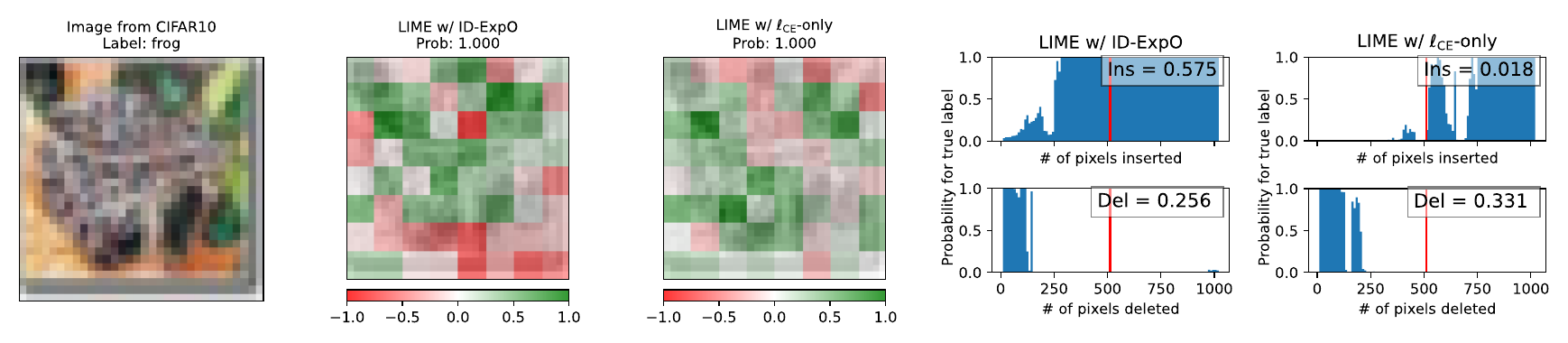} \\
\includegraphics[width=0.85\textwidth]{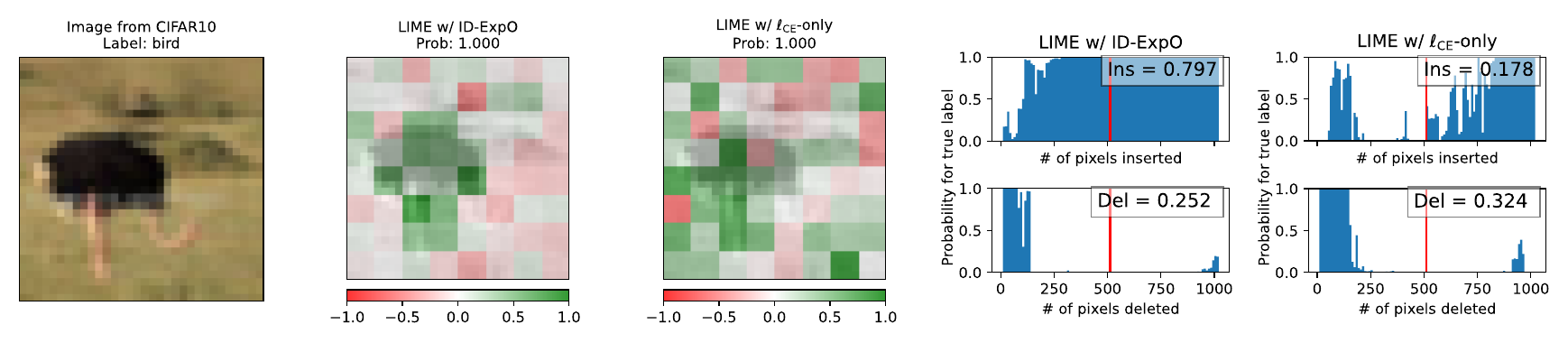} \\
\end{center}

{\bf (B) Grad-CAM on CIFAR-10 ($S=0.5 \cdot HW$)}
\begin{center}
\includegraphics[width=0.85\textwidth]{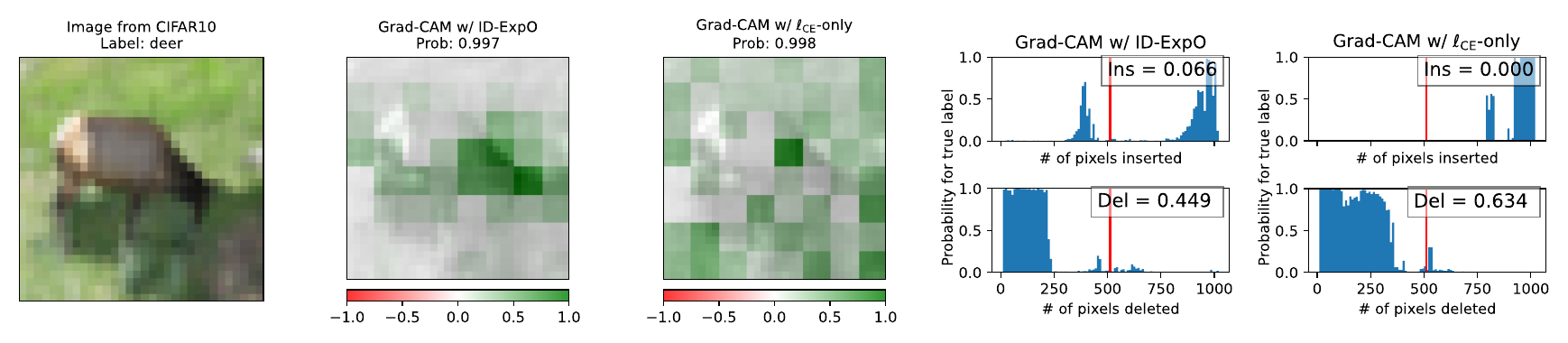} \\
\includegraphics[width=0.85\textwidth]{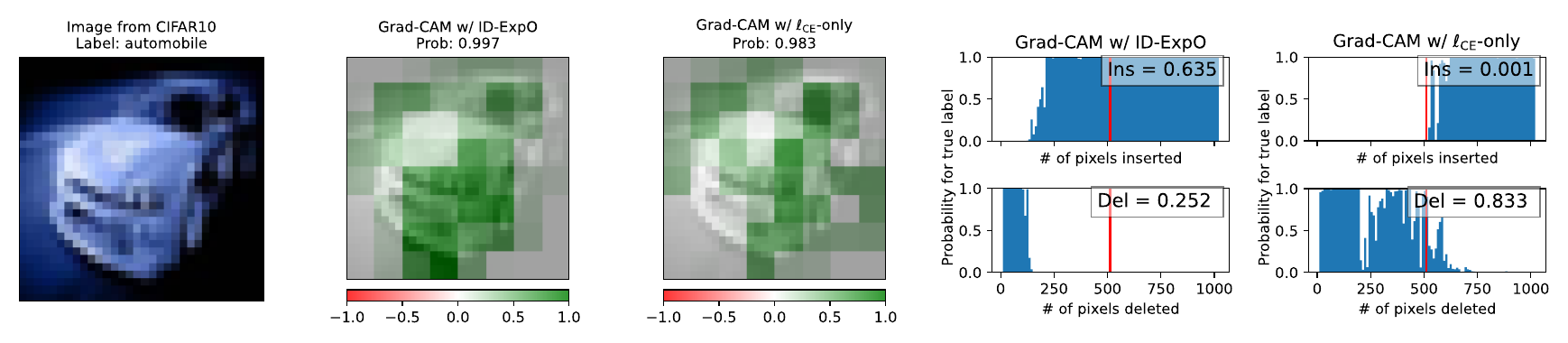} \\
\includegraphics[width=0.85\textwidth]{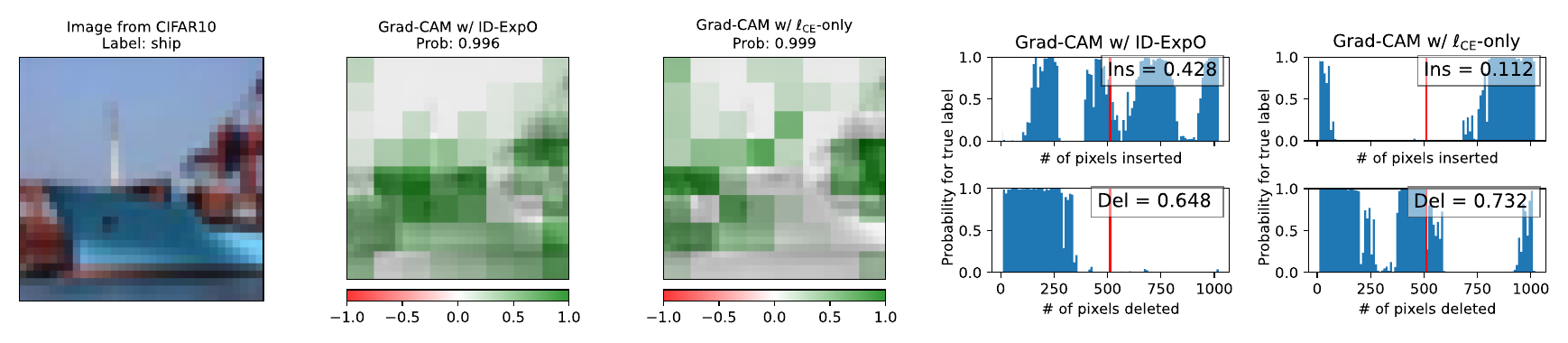} \\
\end{center}
\centering
{\small
\begin{minipage}{0.17\hsize}
\centering
  (a)
\end{minipage}
\begin{minipage}{0.17\hsize}
\centering
  (b)
\end{minipage}
\begin{minipage}{0.17\hsize}
\centering
  (c)
\end{minipage}
\begin{minipage}{0.17\hsize}
\centering
  (d)
\end{minipage}
\begin{minipage}{0.17\hsize}
\centering
  (e)
\end{minipage}
}
\captionsetup{skip=8pt}
\caption{
Examples of the produced explanations on CIFAR-10.
Each row illustrates (a) an input image, (b)--(c) the heatmaps of the explanations by the explainers with ID-ExpO and $\ell_{\mathrm{CE}}$-only, and (d)--(e) the insertion score (top) and the deletion score (bottom) for those explanations in the case of $S=0.5\cdot HW$, which means that the scores are the blue areas to the left of red vertical lines.
}
\label{fig:appendix:image_examples:1}
\end{figure}

\begin{figure}[p!]
\small
{\bf (A) LIME on STL-10 ($S=0.5 \cdot HW$)}
\begin{center}
\includegraphics[width=0.85\textwidth]{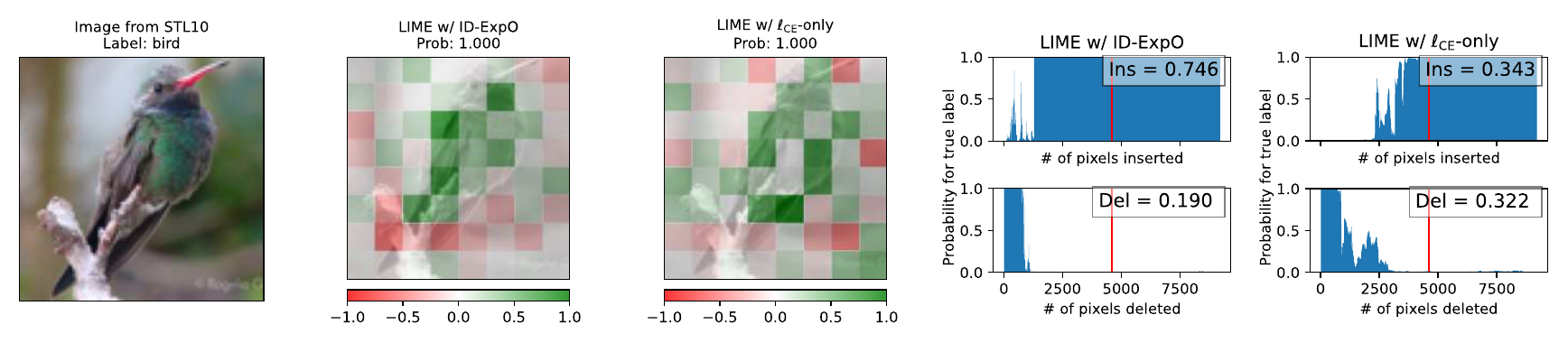} \\
\includegraphics[width=0.85\textwidth]{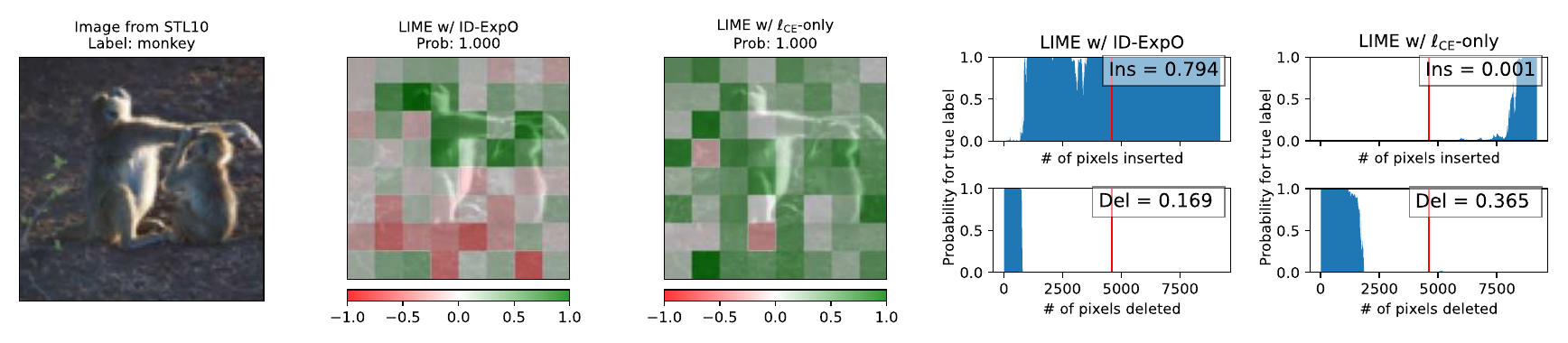} \\
\includegraphics[width=0.85\textwidth]{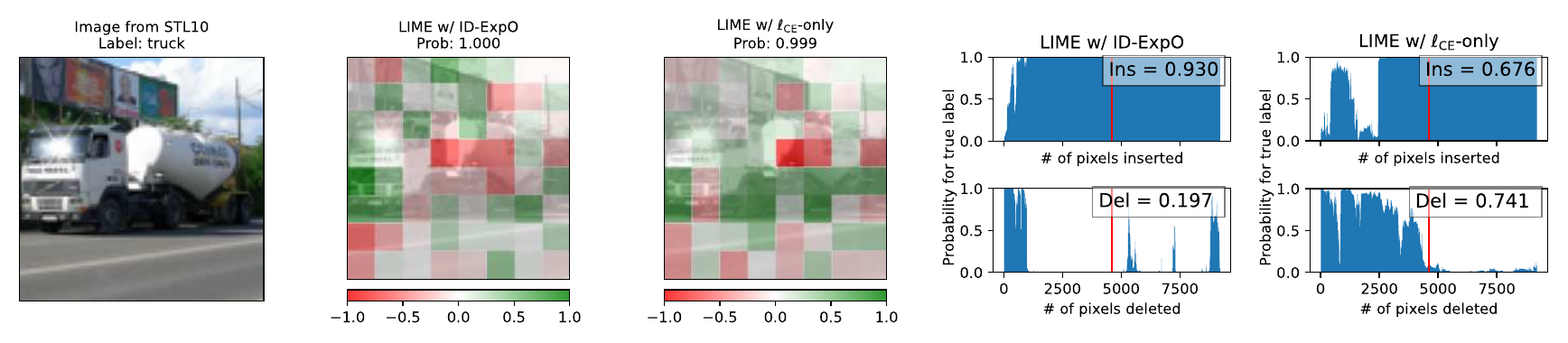} \\
\end{center}

{\bf (B) Grad-CAM on STL-10 ($S=0.5 \cdot HW$)}
\begin{center}
\includegraphics[width=0.85\textwidth]{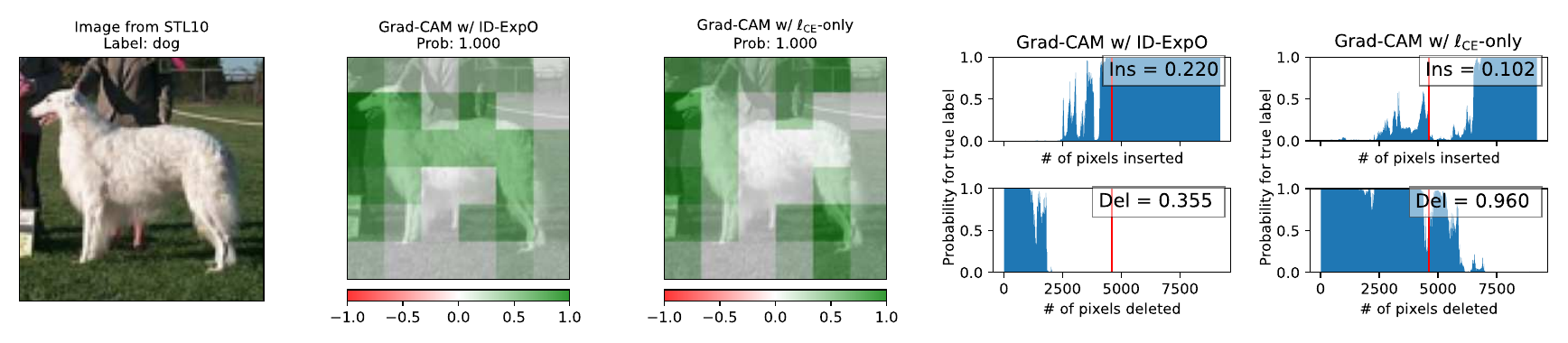} \\
\includegraphics[width=0.85\textwidth]{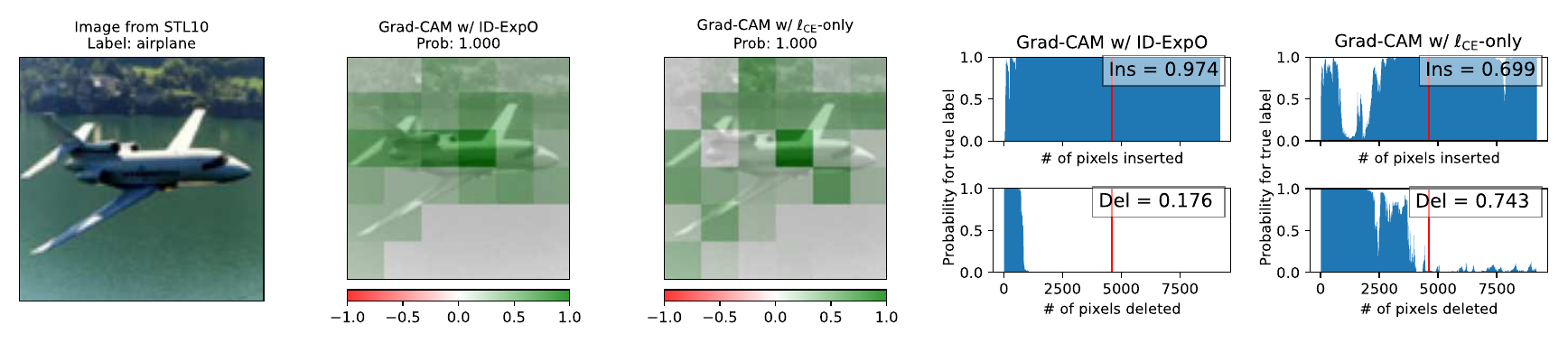} \\
\includegraphics[width=0.85\textwidth]{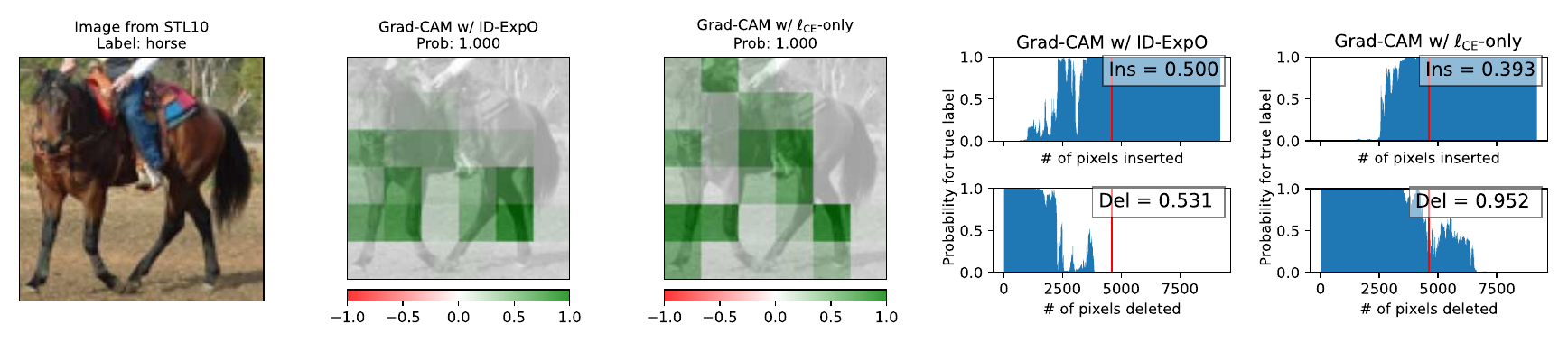} 
\end{center}
\centering
{\small
\begin{minipage}{0.17\hsize}
\centering
  (a)
\end{minipage}
\begin{minipage}{0.17\hsize}
\centering
  (b)
\end{minipage}
\begin{minipage}{0.17\hsize}
\centering
  (c)
\end{minipage}
\begin{minipage}{0.17\hsize}
\centering
  (d)
\end{minipage}
\begin{minipage}{0.17\hsize}
\centering
  (e)
\end{minipage}
}
\captionsetup{skip=8pt}
\caption{
Examples of the produced explanations on STL-10.
How to read the figures is the same as Figure~\ref{fig:appendix:image_examples:1}.
}
\label{fig:appendix:image_examples:2}
\end{figure}

Figures~\ref{fig:appendix:image_examples:1}~and~\ref{fig:appendix:image_examples:2} show additional visualization examples of the produced explanations on image datasets when the insertion and deletion scores of the explanations were improved by using ID-ExpO.
Overall, compared to the explanations of $\ell_{\mathrm{CE}}$-only, the explanations of ID-ExpO had the tendency that large positive contributions were assigned to a part of super-pixels that captures the object of the class label well.

\clearpage
\section{Experiments on Tabular Datasets}\label{sec:appendix:tabular}

\subsection{Tabular Datasets}\label{sec:appendix:tabular:spec}
We used six tabular classification datasets with numerical features from OpenML dataset repository~\parencite{bischl2021openml}: collins, mfeat-fourier, one-hundred-plants-shape, qsar-biodeg, steel-plates-fault, and wine-quality-red.
Table~\ref{tab:appendix:tabular:spec} shows the numbers of samples, features and classes of each tabular dataset.
For each dataset, we created five sets, each of which consists of training, validation and test sets, by randomly dividing the dataset in the ratio of 70\%, 10\% and 20\%.
We standardized each feature value using the training set.

\begin{table}[t]
\centering
\caption{Specification of tabular datasets.}
\label{tab:appendix:tabular:spec}
\begin{tabular}{@{}rrrr@{}}
\toprule
Dataset & \# samples & \# features & \# classes \\ \midrule
collins & 500 & 23 & 2 \\
mfeat-fourier & 2,000 & 77 & 10 \\
one-hundred-plants-shape & 1,600 & 65 & 100 \\
qsar-biodeg & 1,055 & 42 & 2 \\
steel-plates-fault & 1,941 & 28 & 7 \\
wine-quality-red & 1,599 & 12 & 6 \\ \bottomrule
\end{tabular}
\end{table}

\subsection{Implementation Details for Tabular Datasets}\label{sec:appendix:tabular:implementation}

We used a multilayer perceptron (MLP) with two hidden layers of 256 units and ReLU activation functions as a predictor, which was trained on the training set of each dataset in advance in the standard supervised learning manner.
We used LIME and KernelSHAP with the same hyperparameter setting as the LIME for the image classification as explainers.
Since there is no bunch of features like a super-pixel, we directly calculated the contributions of individual features in~\mainornot{\bref{eq:proposed:g_lime}}{(12)}. 
We did not use Grad-CAM because it is impossible to associate the features with the activation maps of the intermediate layers of the MLP.
The optimizer is the same as that for image classification, except we chose the learning rate in the range of $\{0.01, 0.001 \}$. 
The training continued until 200 epochs were reached or the value of $\mathrm{valscore}(f_{\theta};2)$ does not gain for 20 consecutive epochs.

\subsection{Results of Tabular Datasets}\label{sec:appendix:tabular:result}

\begin{figure}[t!]
\centering
\begin{minipage}{0.24\hsize}
\includegraphics[width=\columnwidth]{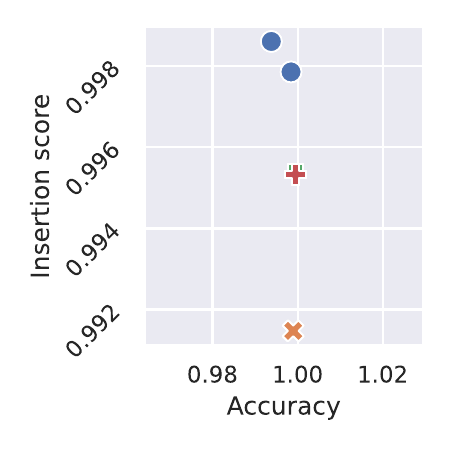}
\end{minipage}
\begin{minipage}{0.24\hsize}
\includegraphics[width=\columnwidth]{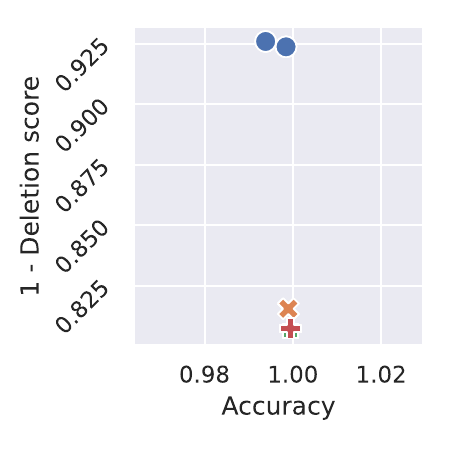} 
\end{minipage}
\begin{minipage}{0.24\hsize}
\includegraphics[width=\columnwidth]{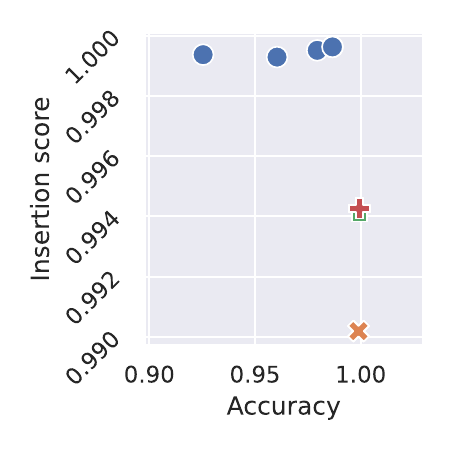}
\end{minipage}
\begin{minipage}{0.24\hsize}
\includegraphics[width=\columnwidth]{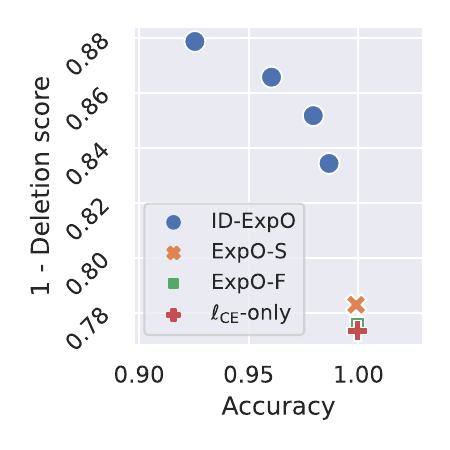} 
\end{minipage}
\caption{
  Mean insertion and mean one-minus-deletion scores against accuracy on steel-plates-fault dataset in case of $S=0.3 \cdot Q$.
  The scores are averaged over the five training/validation/test sets.
  The first two columns show the results on LIME, while the other shows the results on KernelSHAP.
}
\label{fig:experiment:tabular_scores}
\end{figure}

\begin{figure}[t]
\centering
\includegraphics[width=0.4\textwidth]{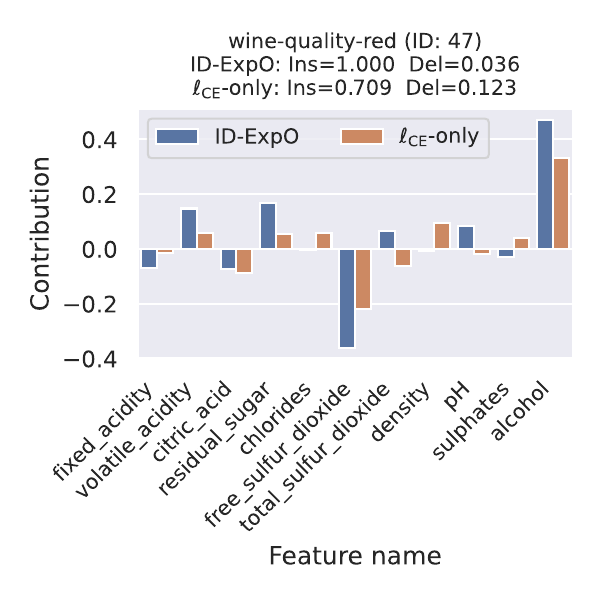}
\caption{Feature contributions on a sample in wine-quality-red dataset.}
\label{fig:experiment:tabular_example}
\end{figure}

Figure~\ref{fig:experiment:tabular_scores} shows the insertion and one-minus-deletion scores against accuracy on steel-plates-fault dataset.
To test the differences among the methods for each evaluation metric, we performed a paired t-test at 5\% level for the results with $\eta=3$.
As a result, in the case of using LIME, ID-ExpO achieved the highest insertion and one-minus-deletion scores, and there was no statistical accuracy difference among the methods.
In the case of using KernelSHAP, although the insertion and one-minus-deletion scores of ID-ExpO were the highest, its accuracy was superior to the comparing methods.
LIME and KernelSHAP are similarly formalized as linear regression models.
The main difference between them is the kernel functions in~\mainornot{\bref{eq:proposed:g_lime}}{(12)}.
Since the Shapley kernel used in KernelSHAP often outputs extremely small values, resulting in unstable calculation in~\mainornot{\bref{eq:proposed:g_lime}}{(12)}, it may have adversely affected the accuracy.
As shown in Appendix~\ref{sec:appendix:tabular:scores}, similar results were observed on the other tabular datasets.

To analyze how the explanations changed by employing ID-ExpO, we visualize the feature contributions by LIME with ID-ExpO and $\ell_{\mathrm{CE}}$-only.
We show a typical example in the wine-quality-red dataset in Figure~\ref{fig:experiment:tabular_example}.
In many samples, including this, we found that ID-ExpO tended to bring larger positive or negative feature contributions than $\ell_{\mathrm{CE}}$-only, although LIME for both has the same setting.
This is because that ID-ExpO adjusts the predictor's behaviors so that important features are taken into account early in the calculation of the insertion and deletion metrics.
Since the explanation with such feature contributions makes features that users should focus on more clear, ID-ExpO can be effective in producing explanations that are easy to understand for the users.

\subsection{All Quantitative Results on Tabular Datasets}\label{sec:appendix:tabular:scores}
\begin{figure}[p!]
\small
{\bf (A) LIME and KernelSHAP on collins ($S=0.3\cdot Q$)}\\
\begin{minipage}{0.245\hsize}
\includegraphics[width=\textwidth]{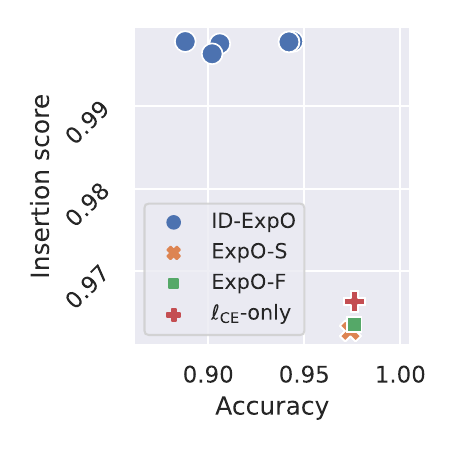}
\end{minipage}
\begin{minipage}{0.245\hsize}
\includegraphics[width=\textwidth]{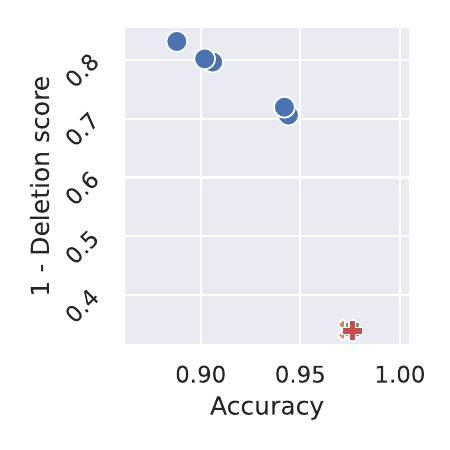} 
\end{minipage}
\begin{minipage}{0.245\hsize}
\includegraphics[width=\textwidth]{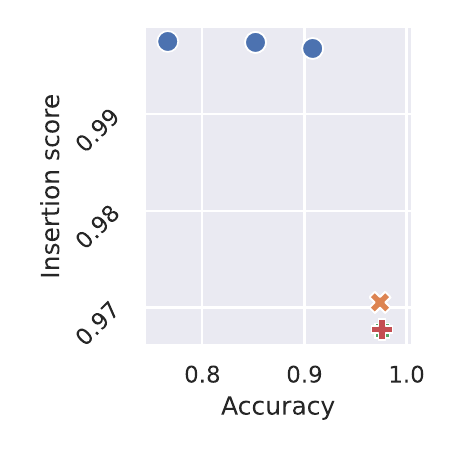}
\end{minipage}
\begin{minipage}{0.245\hsize}
\includegraphics[width=\textwidth]{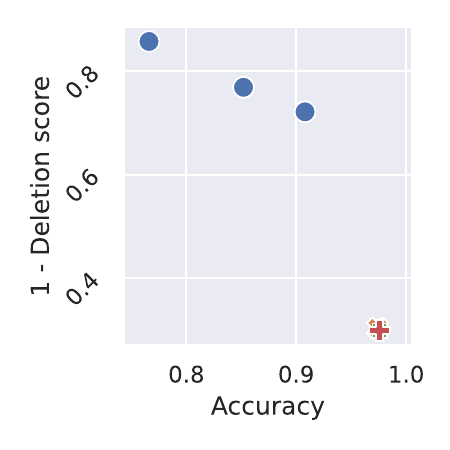} 
\end{minipage}
\\
{\bf (B) LIME and KernelSHAP on collins ($S=0.5\cdot Q$)}\\
\begin{minipage}{0.245\hsize}
\includegraphics[width=\textwidth]{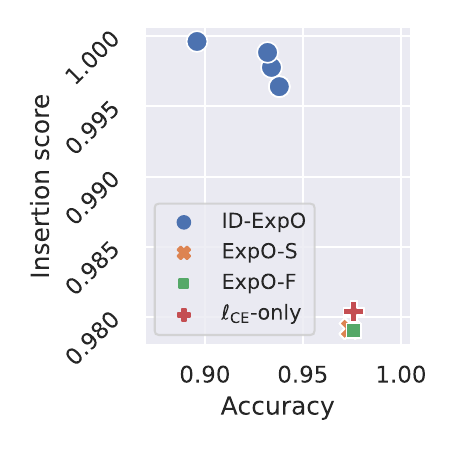}
\end{minipage}
\begin{minipage}{0.245\hsize}
\includegraphics[width=\textwidth]{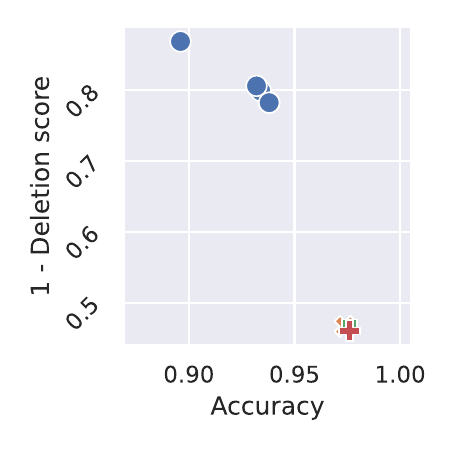} 
\end{minipage}
\begin{minipage}{0.245\hsize}
\includegraphics[width=\textwidth]{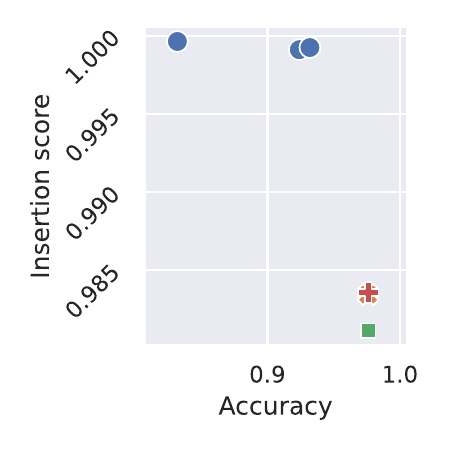}
\end{minipage}
\begin{minipage}{0.245\hsize}
\includegraphics[width=\textwidth]{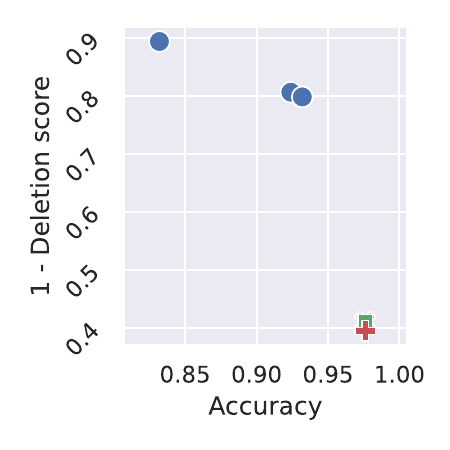} 
\end{minipage}
\\
{\bf (C) LIME and KernelSHAP on mfeat-fourier ($S=0.3\cdot Q$)}\\
\begin{minipage}{0.245\hsize}
\includegraphics[width=\textwidth]{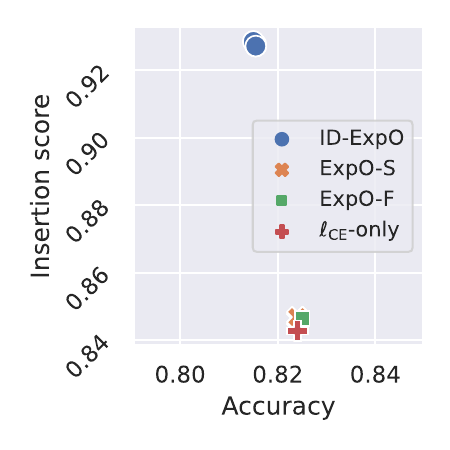}
\end{minipage}
\begin{minipage}{0.245\hsize}
\includegraphics[width=\textwidth]{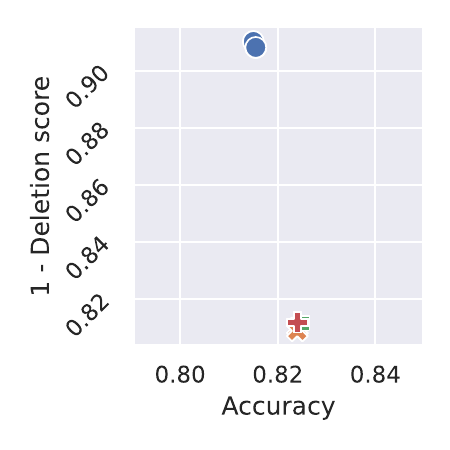} 
\end{minipage}
\begin{minipage}{0.245\hsize}
\includegraphics[width=\textwidth]{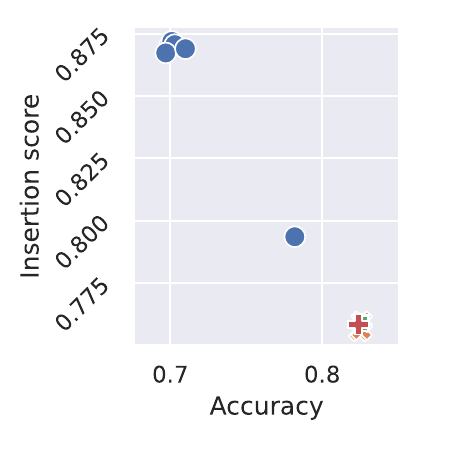}
\end{minipage}
\begin{minipage}{0.245\hsize}
\includegraphics[width=\textwidth]{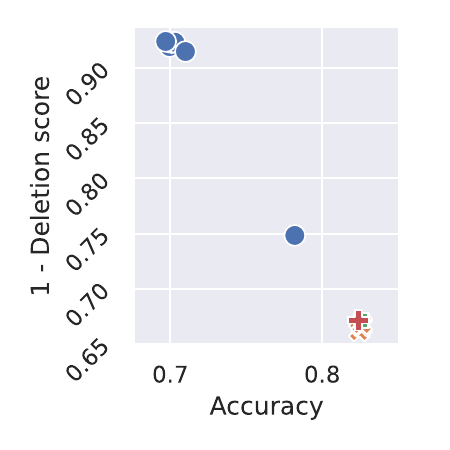} 
\end{minipage}
\\
{\bf (D) LIME and KernelSHAP on mfeat-fourier ($S=0.5\cdot Q$)}\\
\begin{minipage}{0.245\hsize}
\includegraphics[width=\textwidth]{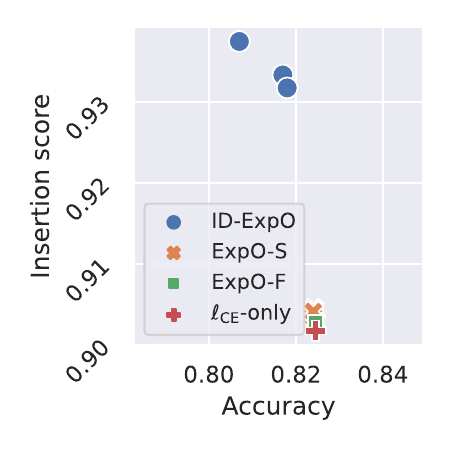}
\end{minipage}
\begin{minipage}{0.245\hsize}
\includegraphics[width=\textwidth]{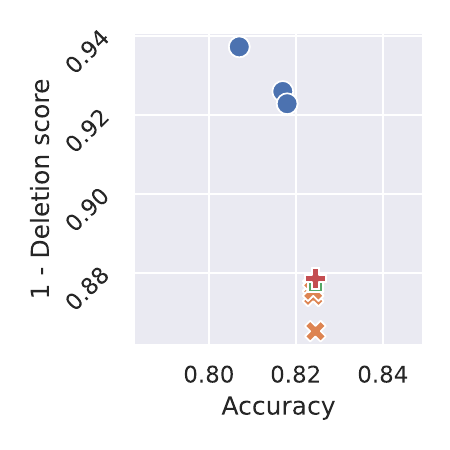} 
\end{minipage}
\begin{minipage}{0.245\hsize}
\includegraphics[width=\textwidth]{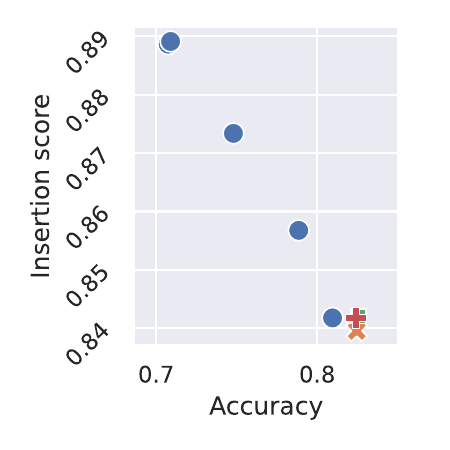}
\end{minipage}
\begin{minipage}{0.245\hsize}
\includegraphics[width=\textwidth]{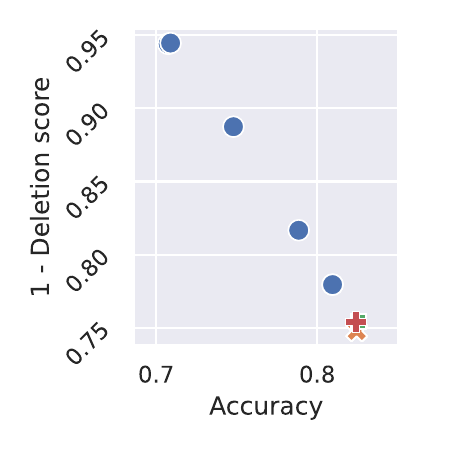} 
\end{minipage}
\caption{
  Mean insertion and mean one-minus-deletion scores against accuracy on collins and mfeat-fourier datasets.
  The scores are averaged over the five training/validation/test sets.
  The first two columns show the results on LIME, while the others show the results on KernelSHAP.
  Each point has a different accuracy weight $\eta \in \{0.5, 1.0. \cdots, 3.0 \}$.
  The higher the score, the better.
}
\label{fig:appendix:tabular_scores:1}
\end{figure}

\begin{figure}[p!]
\small
{\bf (A) LIME and KernelSHAP on one-hundred-plants-shape ($S=0.3\cdot Q$)}\\
\begin{minipage}{0.245\hsize}
\includegraphics[width=\textwidth]{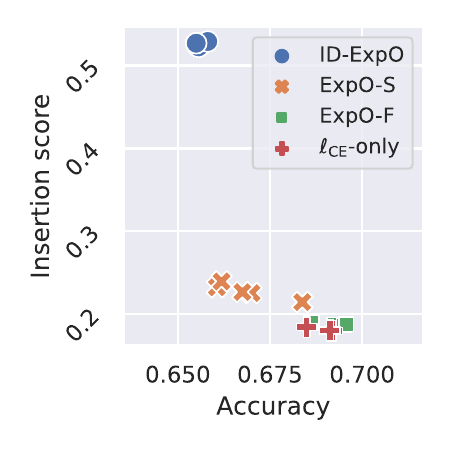}
\end{minipage}
\begin{minipage}{0.245\hsize}
\includegraphics[width=\textwidth]{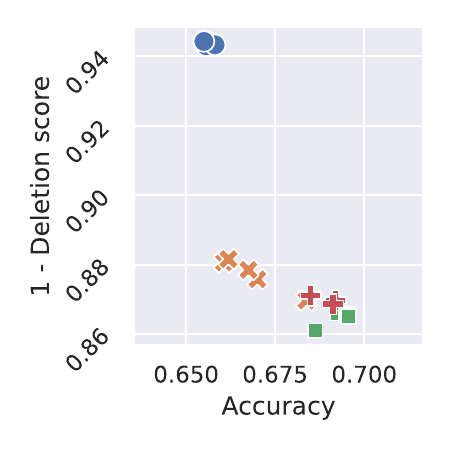} 
\end{minipage}
\begin{minipage}{0.245\hsize}
\includegraphics[width=\textwidth]{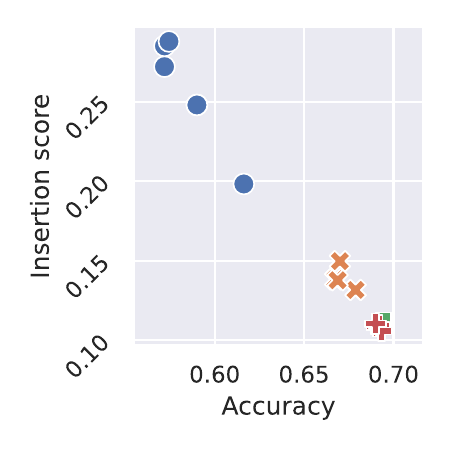}
\end{minipage}
\begin{minipage}{0.245\hsize}
\includegraphics[width=\textwidth]{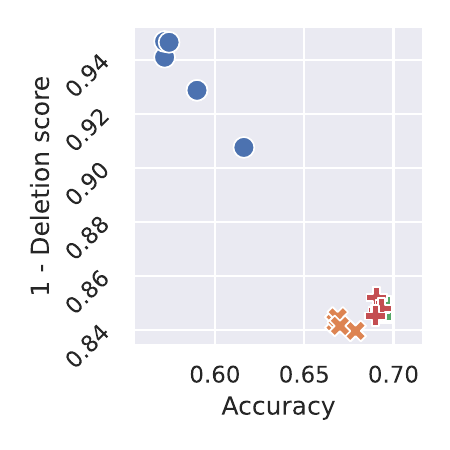} 
\end{minipage}
\\
{\bf (B) LIME and KernelSHAP on one-hundred-plants-shape ($S=0.5\cdot Q$)}\\
\begin{minipage}{0.245\hsize}
\includegraphics[width=\textwidth]{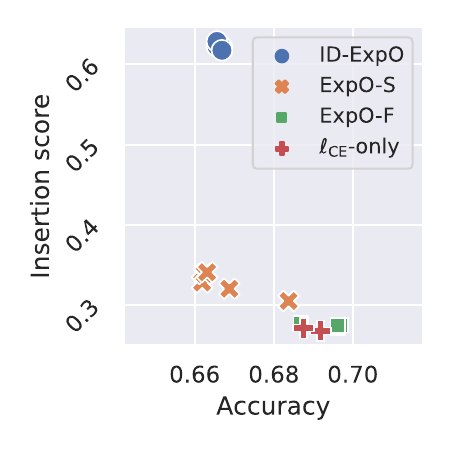}
\end{minipage}
\begin{minipage}{0.245\hsize}
\includegraphics[width=\textwidth]{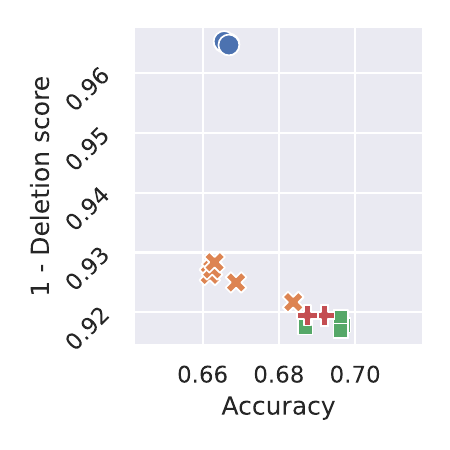} 
\end{minipage}
\begin{minipage}{0.245\hsize}
\includegraphics[width=\textwidth]{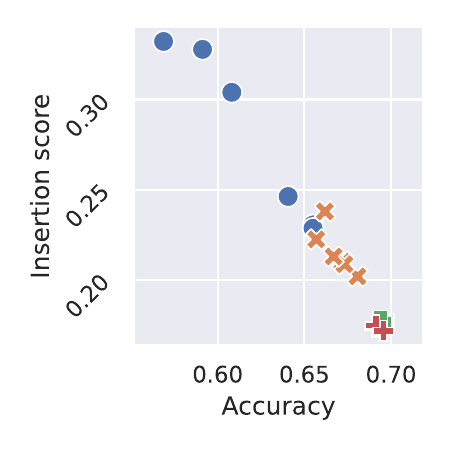}
\end{minipage}
\begin{minipage}{0.245\hsize}
\includegraphics[width=\textwidth]{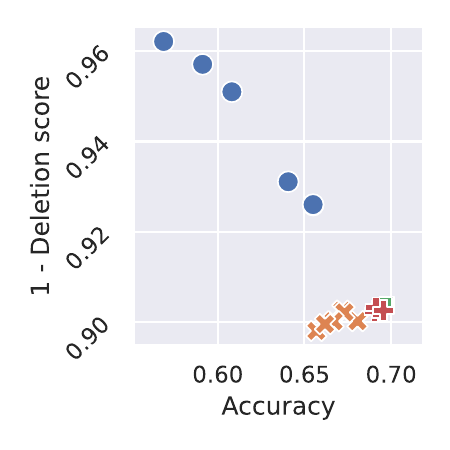} 
\end{minipage}
\\
{\bf (C) LIME and KernelSHAP on qsar-biodeg ($S=0.3\cdot Q$)}\\
\begin{minipage}{0.245\hsize}
\includegraphics[width=\textwidth]{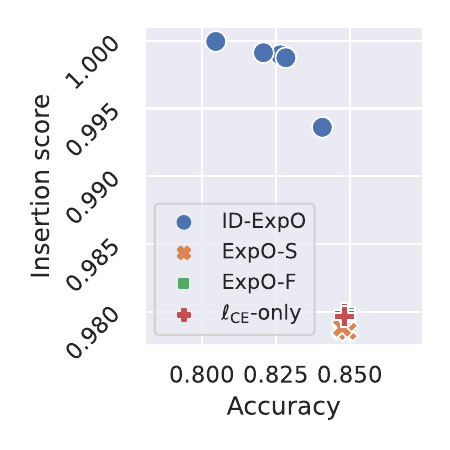}
\end{minipage}
\begin{minipage}{0.245\hsize}
\includegraphics[width=\textwidth]{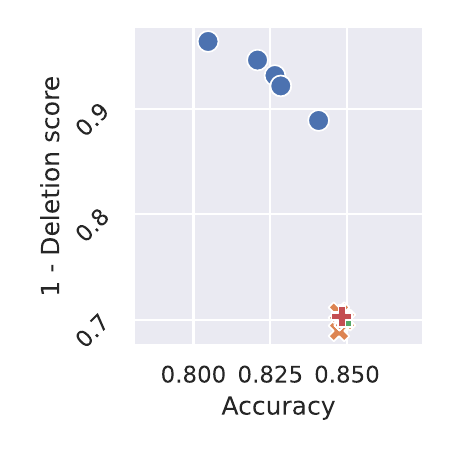} 
\end{minipage}
\begin{minipage}{0.245\hsize}
\includegraphics[width=\textwidth]{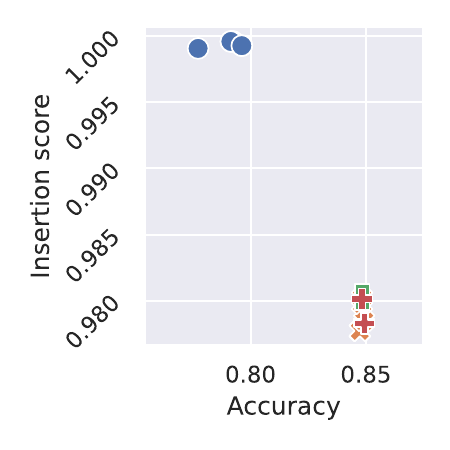}
\end{minipage}
\begin{minipage}{0.245\hsize}
\includegraphics[width=\textwidth]{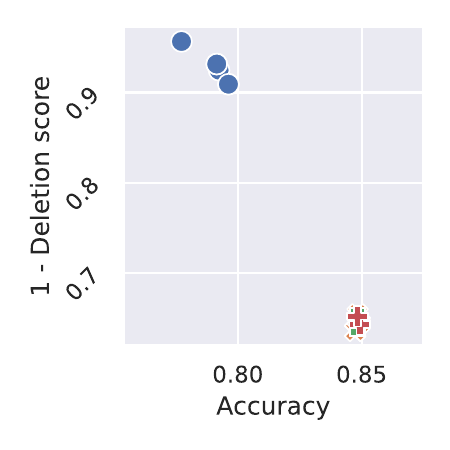} 
\end{minipage}
\\
{\bf (D) LIME and KernelSHAP on qsar-biodeg ($S=0.5\cdot Q$)}\\
\begin{minipage}{0.245\hsize}
\includegraphics[width=\textwidth]{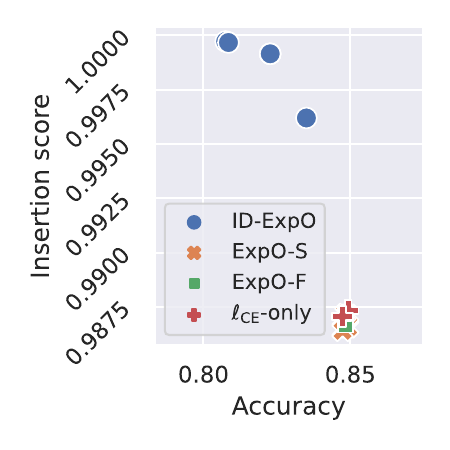}
\end{minipage}
\begin{minipage}{0.245\hsize}
\includegraphics[width=\textwidth]{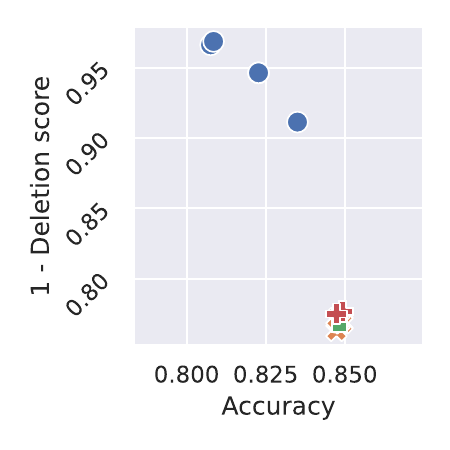} 
\end{minipage}
\begin{minipage}{0.245\hsize}
\includegraphics[width=\textwidth]{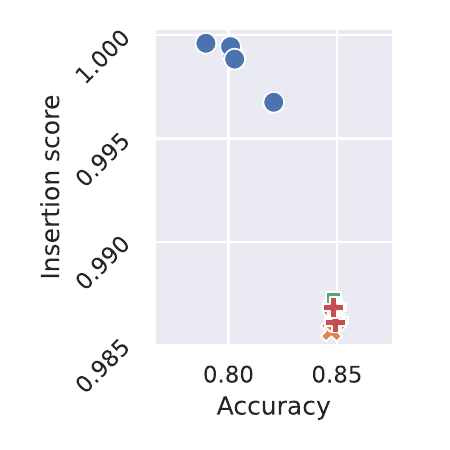}
\end{minipage}
\begin{minipage}{0.245\hsize}
\includegraphics[width=\textwidth]{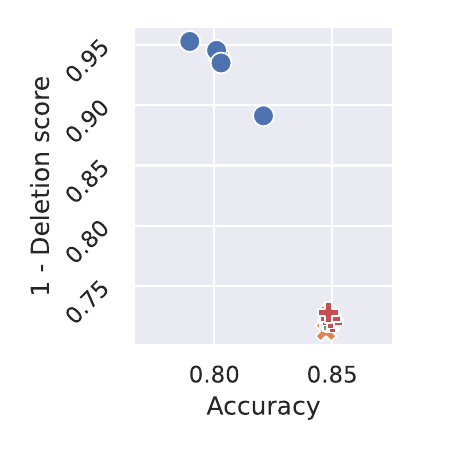} 
\end{minipage}
\caption{
  Mean insertion and mean one-minus-deletion scores against accuracy on one-hundred-plants-shape and qsar-biodeg datasets.
  How to read these figures is the same as Figure~\ref{fig:appendix:tabular_scores:1}.
}
\label{fig:appendix:tabular_scores:2}
\end{figure}

\begin{figure}[p!]
\small
{\bf (A) LIME and KernelSHAP on steel-plates-fault ($S=0.3\cdot Q$)}\\
\begin{minipage}{0.245\hsize}
\includegraphics[width=\textwidth]{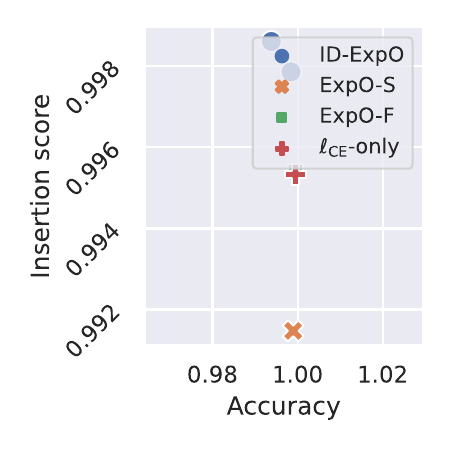}
\end{minipage}
\begin{minipage}{0.245\hsize}
\includegraphics[width=\textwidth]{fig/tabular_scatter_outlined/scatter_steel-plates-fault_LIME_K0.3_acc_del.pdf} 
\end{minipage}
\begin{minipage}{0.245\hsize}
\includegraphics[width=\textwidth]{fig/tabular_scatter_outlined/scatter_steel-plates-fault_SHAP_K0.3_acc_ins.pdf}
\end{minipage}
\begin{minipage}{0.245\hsize}
\includegraphics[width=\textwidth]{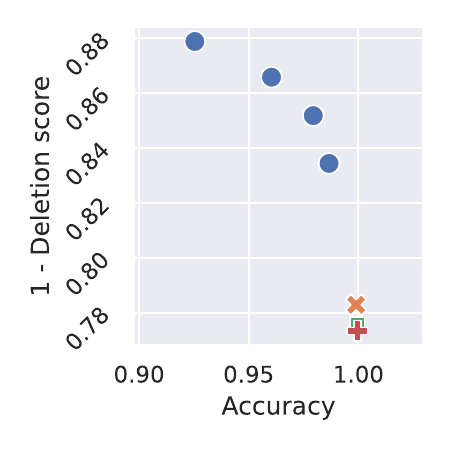} 
\end{minipage}
\\
{\bf (B) LIME and KernelSHAP on steel-plates-fault ($S=0.5\cdot Q$)}\\
\begin{minipage}{0.245\hsize}
\includegraphics[width=\textwidth]{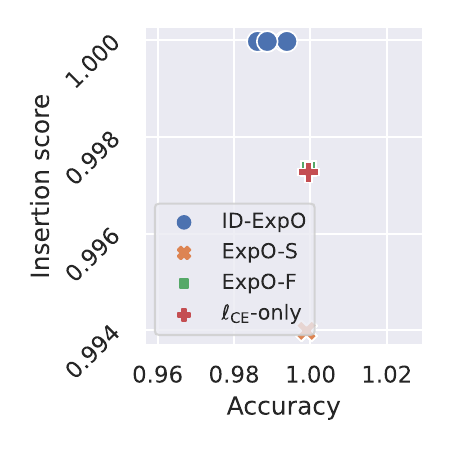}
\end{minipage}
\begin{minipage}{0.245\hsize}
\includegraphics[width=\textwidth]{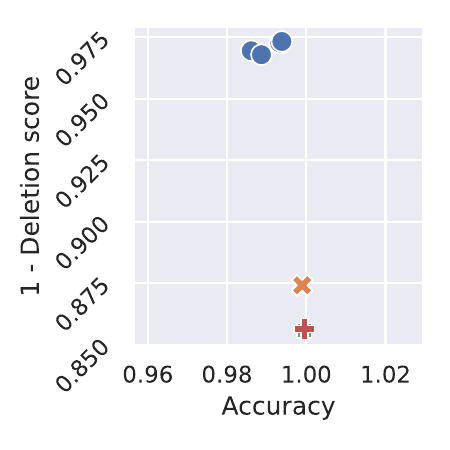} 
\end{minipage}
\begin{minipage}{0.245\hsize}
\includegraphics[width=\textwidth]{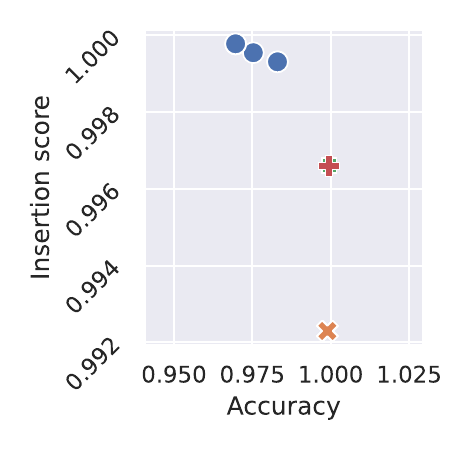}
\end{minipage}
\begin{minipage}{0.245\hsize}
\includegraphics[width=\textwidth]{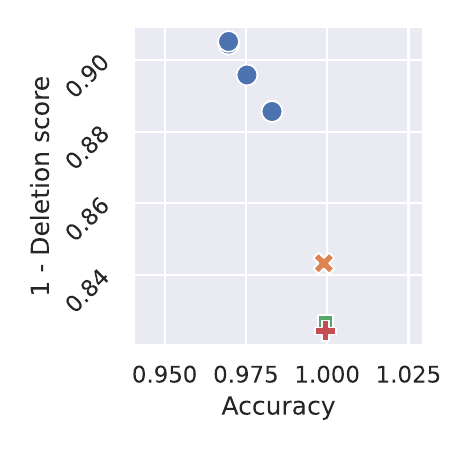} 
\end{minipage}
\\
{\bf (C) LIME and KernelSHAP on wine-quality-red ($S=0.3\cdot Q$)}\\
\begin{minipage}{0.245\hsize}
\includegraphics[width=\textwidth]{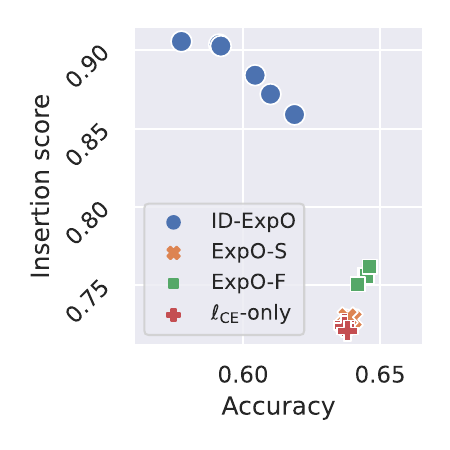}
\end{minipage}
\begin{minipage}{0.245\hsize}
\includegraphics[width=\textwidth]{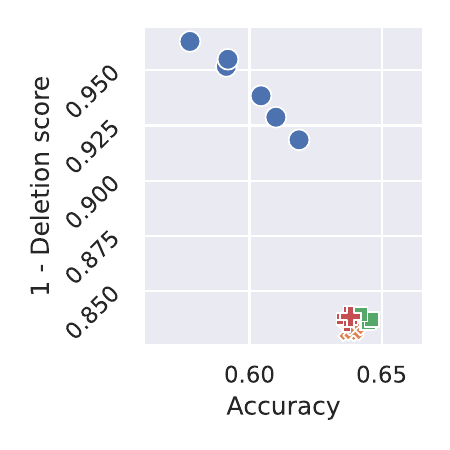} 
\end{minipage}
\begin{minipage}{0.245\hsize}
\includegraphics[width=\textwidth]{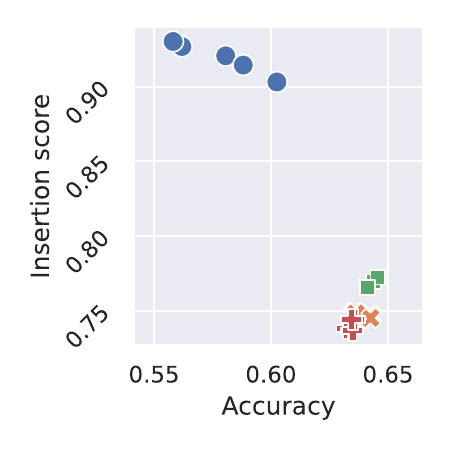}
\end{minipage}
\begin{minipage}{0.245\hsize}
\includegraphics[width=\textwidth]{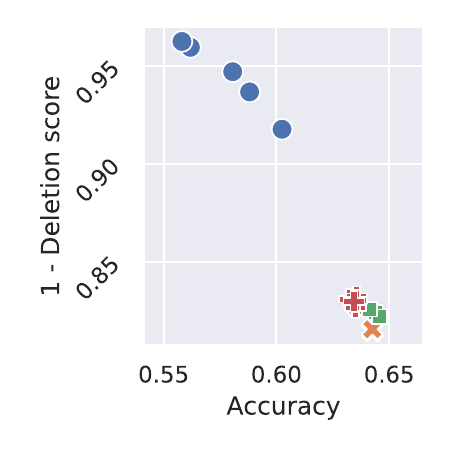} 
\end{minipage}
\\
{\bf (D) LIME and KernelSHAP on wine-quality-red ($S=0.5\cdot Q$)}\\
\begin{minipage}{0.245\hsize}
\includegraphics[width=\textwidth]{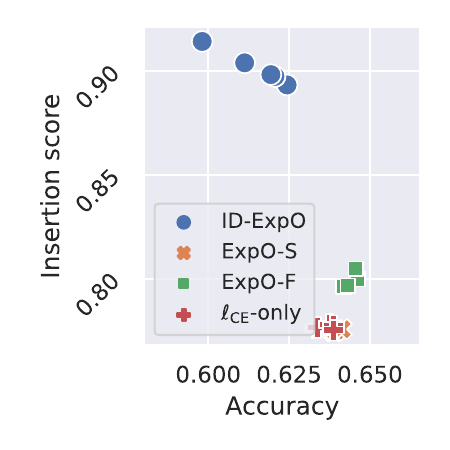}
\end{minipage}
\begin{minipage}{0.245\hsize}
\includegraphics[width=\textwidth]{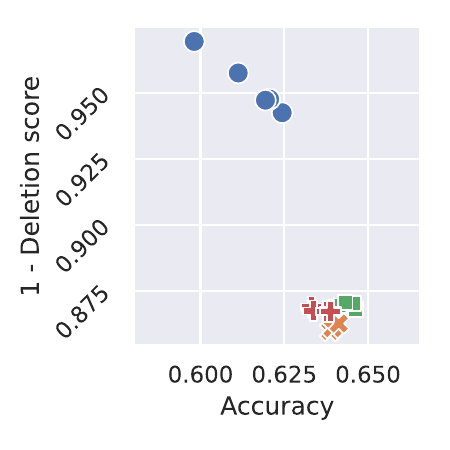} 
\end{minipage}
\begin{minipage}{0.245\hsize}
\includegraphics[width=\textwidth]{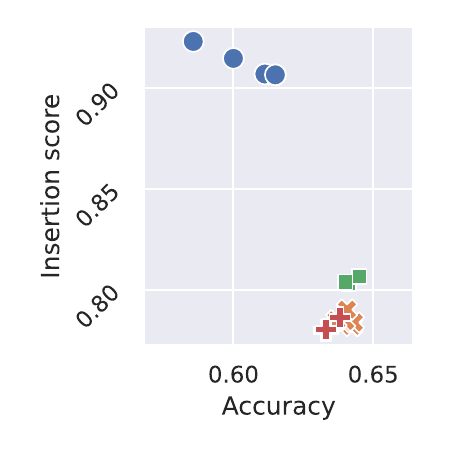}
\end{minipage}
\begin{minipage}{0.245\hsize}
\includegraphics[width=\textwidth]{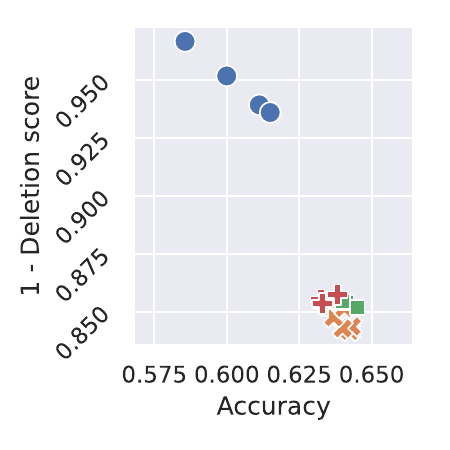} 
\end{minipage}
\caption{
  Mean insertion and mean one-minus-deletion scores against accuracy on steel-plates-fault and wine-quality-red datasets.
  How to read these figures is the same as Figure~\ref{fig:appendix:tabular_scores:1}.
}
\label{fig:appendix:tabular_scores:3}
\end{figure}

Figures~\ref{fig:appendix:tabular_scores:1}--\ref{fig:appendix:tabular_scores:3} show the insertion and the one-minus-deletion scores against the accuracy on the six tabular datasets.
On all the datasets, ID-ExpO outperformed the others in terms of the insertion and the one-minus-deletion scores for any in the range of $\eta$.
With the accuracy, by putting emphasis on the accuracy, i.e., by setting $\eta = 3$, we found that ID-ExpO could keep comparable or slightly low accuracy compared to the others, while the highest insertion and one-minus-deletion scores.
However, in some cases of using KernelSHAP, e.g., Figures~\ref{fig:appendix:tabular_scores:2}(C)~and~\ref{fig:appendix:tabular_scores:3}(C), we found that the accuracy of ID-ExpO degraded compared to the other methods.

\clearpage
\section{Broader Impact}\label{sec:appendix:impacts}
The proposed method can contribute to producing faithful explanations that capture the predictor's behaviors well.
However, the fact does not guarantee that the explanations are easy-to-understand for humans.
If the predictor and explainer that are trained using the proposed method produce explanations that are faithful but difficult to understand for humans, they might give users wrong interpretations of the prediction results.
There are several studies that explore producing explanations that are easy to understand for humans, which include human-in-the-loop approaches~\parencite{Lage2018-tn,gao2022aligning} and using ground truths of explanations by human annotators~\parencite{ross2017right,balayan2020teaching}.
By using the proposed method together with such approaches, we can alleviate the concern of such misinterpretation while keeping high faithfulness in the explanations.

\section{Limitations}\label{sec:appendix:limitations}
The proposed method can contribute to producing faithful explanations that capture the predictor's behaviors well.
However, the fact does not guarantee that the explanations are easy-to-understand for humans.
There are several studies that explore producing explanations that are easy to understand for humans, which include human-in-the-loop approaches~\parencite{Lage2018-tn,gao2022aligning} and using the ground truth of explanations by human annotators~\parencite{ross2017right,balayan2020teaching}.
When we require easy-to-understand explanations, combining the proposed method with such approaches would result in producing faithful and easy-to-understand explanations.

The proposed method is applicable to a wide range of predictors and explainers.
In practice, the computational complexities of the predictor and explainer we use can be barriers to using the proposed method. 
The perturbation-based explainers, such as LIME and KernelSHAP, produce an explanation for an input sample by using $M$ perturbed samples around the input sample.
In our experiment, $M$ was set to 200, and the mini-batch size was 128. 
This means that $200 \times 128 = 25,600$ samples were used to update the predictor's parameters $\theta$ once.
For this issue, fast computation is possible by performing the data parallel training using multiple GPUs.
On the other hand, using Grad-CAM, one of the gradient-based explainers, as an explainer in the proposed method is computationally more efficient than LIME and KernelSHAP because it does not need to increase the sample.
Note that, although those computational complexities affect training efficiency using the proposed method, the computational complexities of the predictor and explainer in testing are invariant before and after applying the proposed method.

\ifismaindoc\else\printbibliography\fi

\end{document}

\end{document}